\documentclass[10pt,twocolumn,letterpaper]{article}

\usepackage[table, dvipsnames]{xcolor}
\usepackage{cvpr}              
\usepackage{bbm}
\usepackage{indentfirst}
\usepackage{graphicx}
\usepackage{marvosym}

\definecolor{cvprblue}{rgb}{0.21,0.49,0.74}
\usepackage[pagebackref,breaklinks,colorlinks,allcolors=cvprblue]{hyperref}

\title{Multi-focal Conditioned Latent Diffusion for Person Image Synthesis}

\author{Jiaqi Liu$^{1}$  \quad Jichao Zhang$^{2}$\href{mailto:zhang163220@gmail.com}{\textsuperscript{\Letter}}  \quad Paolo Rota$^{1}$  \quad Nicu Sebe$^{1}$ \\
University of Trento$^{1}$ \quad \quad Ocean University of China$^{2}$}

\begin{document}
\maketitle
\begin{abstract}
The Latent Diffusion Model (LDM) has demonstrated strong capabilities in high-resolution image generation and has been widely employed for Pose-Guided Person Image Synthesis (PGPIS), yielding promising results. However, the compression process of LDM often results in the deterioration of details, particularly in sensitive areas such as facial features and clothing textures. In this paper, we propose a Multi-focal Conditioned Latent Diffusion (MCLD) method to address these limitations by conditioning the model on disentangled, pose-invariant features from these sensitive regions. Our approach utilizes a multi-focal condition aggregation module, which effectively integrates facial identity and texture-specific information, enhancing the model’s ability to produce appearance realistic and identity-consistent images. Our method demonstrates consistent identity and appearance generation on the DeepFashion dataset and enables flexible person image editing due to its generation consistency. The code is available at \textcolor{magenta}{\url{https://github.com/jqliu09/mcld}}.

\end{abstract}    
\section{Introduction}
\label{sec:intro}

The pose-guided person image synthesis (PGPIS) task focuses on transforming a source image of a person into a target pose, while preserving the appearance and identity of the individual as accurately as possible. This task has significant implications in applications like virtual reality, e-commerce, and the fashion industry, where maintaining photorealistic quality and identity consistency is essential.

Recent approaches to PGPIS largely rely on Generative Adversarial Networks (GANs)~\cite{goodfellow2014generative}, which, despite their success, often struggle with training instability and mode collapse, resulting in suboptimal preservation of identity and garment details~\cite{zhu2019progressive,zhou2021cocosnet,tang2020bipartite,zhang2021pise,men2020controllable,ren2022neural}. As an alternative, diffusion models~\cite{rombach2022high,ho2020denoising} have shown promise in generating high-quality images by progressively refining details through multiple denoising steps. The introduction of PIDM~\cite{bhunia2023person} marked the first application of diffusion models for PGPIS, where latent diffusion models (LDM)~\cite{rombach2022high} compress images into high-level feature representations, thereby reducing the computational complexity while supporting high-resolution outputs. Extensions such as PoCoLD~\cite{han2023controllable} enhance 3D pose correspondence using pose-constrained attention, and CFLD~\cite{lu2024coarse} emphasize semantic understanding with hybrid-granularity attention. 

\begin{figure}[!t]
\centering
\includegraphics[width=1\linewidth]{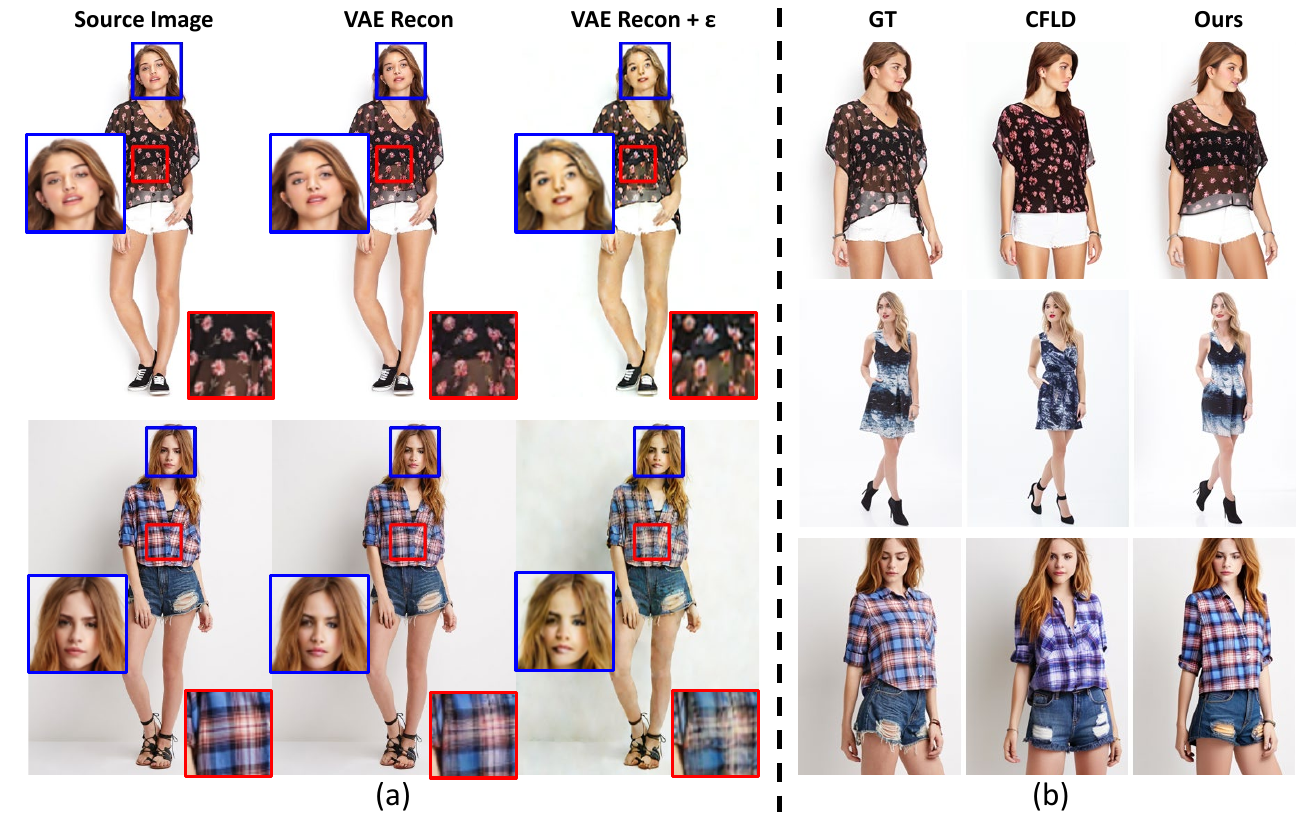}
\caption{(a) The VAE~\cite{rombach2022high} reconstruction deteriorates the detailed information of person images, especially the facial regions and complex textures. These issues worsen for the generated latent with small deviations. A small deviation $\epsilon=0.2$ is added to demonstrate the often case of generated latent. (b) Our methods preserve this detailed information better than other LDM-based methods by introducing multi-focal conditions.}
\label{fig::teaser}
\end{figure}

Despite these advancements, LDM-based methods encounter limitations in recovering fine appearance details, especially in facial and clothing regions. As shown in Fig.~\ref{fig::teaser} (a), this challenge is primarily due to the lossy nature of autoencoder compression~\cite{avrahami2023blended}, which can degrade complex textures and identity-specific features during encoding. Since the lossy reconstructed images are the upper bound of generated images of LDM-based methods, this issue worsens when doing inference since the generated latent deviates from the compressed real latent. Additionally, LDM’s reliance on whole-image conditioning often struggles to focus on sensitive regions where appearance precision is critical. The integration of pose and appearance information complicates detail reconstruction, leading to suboptimal performance across diverse poses and sensitive areas.

To overcome these limitations, we introduce a \textbf{M}ulti-focal \textbf{C}onditioned \textbf{L}atent \textbf{D}iffusion (MCLD) approach for PGPIS. Our method mitigates the loss of detail in sensitive regions by conditioning the diffusion model on the corresponding selectively decoupled features rather than the entire image. Specifically, we isolate high-frequency regions, such as facial identity and appearance textures, from the source image and treat them as independent conditions. This decoupling strategy enhances control over sensitive areas, ensuring better identity preservation and texture fidelity. Our approach first generates pose-invariant embeddings of the selected regions shared in the source and target images using pretrained modules, which are then fused within the Multi-focal Condition Aggregation module. This module introduces selective cross-attention layers, leveraging the structural advantages of UNet to combine the conditions effectively. Consequently, our MCLD method achieves improved control and accuracy, facilitating high-quality, realistic person image synthesis.
Our main contributions can be summarized as follows:
\begin{itemize}
    \item We introduce a new approach, MCLD, that focuses on alleviating the deterioration of important details in sensitive areas like the face and clothing by using separate conditions for these regions, which improves both identity preservation and appearance fidelity.
    \item We develop a multi-focal condition aggregation module that combines controls from multiple focus areas, allowing our model to produce more realistic images without losing or collapse of details in key regions.
    \item Our method achieves consistent appearance generation across different poses, especially in challenging regions like faces and textures, leading to state-of-the-art results on the Deepfashion dataset~\cite{liu2016deepfashion} and flexible-but-accurate editing downstream applications. 
\end{itemize}

\section{Related Works}

\noindent \textbf{Pose-Guided Person Image Synthesis} was initially proposed by PG2~\cite{ma2017pose}, which firstly applied conditional GANs to adversarially refine pose-guided human generation. 
Later, GAN-based research addressed this problem through two main approaches. The first focuses on the transfer process, where methods model the deformation between poses using affine transformations~\cite{siarohin2018deformable, siarohin2021PAMI} and flow fields~\cite{ren2020deep,ren2021flow,ren2021combining,li2019dense}. The second approach aims to enhance the generation quality by better disentangling pose and appearance information. This disentanglement can be implicitly achieved by modeling the spatial correspondence between the pose and appearance features~\cite{zhu2019progressive,zhou2021cocosnet,tang2020bipartite,zhang2021pise,men2020controllable,ren2022neural, zhou2022cross}. Auxiliary explicit information is also introduced to improve the appearance quality, especially for UV texture map~\cite{sarkar2021style, sarkar2020neural, grigorev2019coordinate, zablotskaia2019dwnet} that provides pose-irrelevant appearance guidance. However, due to the instability in training and the mode collapse issues associated with GAN models, previous GAN-based works have encountered challenges with unrealistic or changed textures in posed person images.

To mitigate this limitation, diffusion based methods have been more recently introduced in PGPIS. PIDM~\cite{bhunia2023person} was the first to utilize the iterative denoising property of the diffusion model. Subsequent methods~\cite{han2023controllable,lu2024coarse} have sought to improve the generation capability by employing latent diffusion models~\cite{rombach2022high} (LDM) rather than the pixel space. In detail, CFLD~\cite{lu2024coarse} addresses the importance of semantic understanding towards the decoupling of fine-grained appearance and poses while PoCoLD~\cite{han2023controllable} establishes the correspondence between pose and appearance. More recent some video person animation methods also took the benefit of compressed latent in LDM, but they mainly concentrated on keeping the temporal consistency by spatial attention~\cite{hu2024animate, xu2024magicanimate} and consistent pose guidance~\cite{zhu2024champ}. Both the image and video synthesis methods use a source person image as condition and the generated image would collapse when the target pose varies greatly from the source image. In addition, it has been noticed that there is a deterioration~\cite{han2023controllable, avrahami2023blended} of image quality when LDM compresses images to lower dimensions, especially for images of high-frequency information. However, very few considered tackling this problem. 



\begin{figure*}[!t]
\centering
\includegraphics[width=1\linewidth]{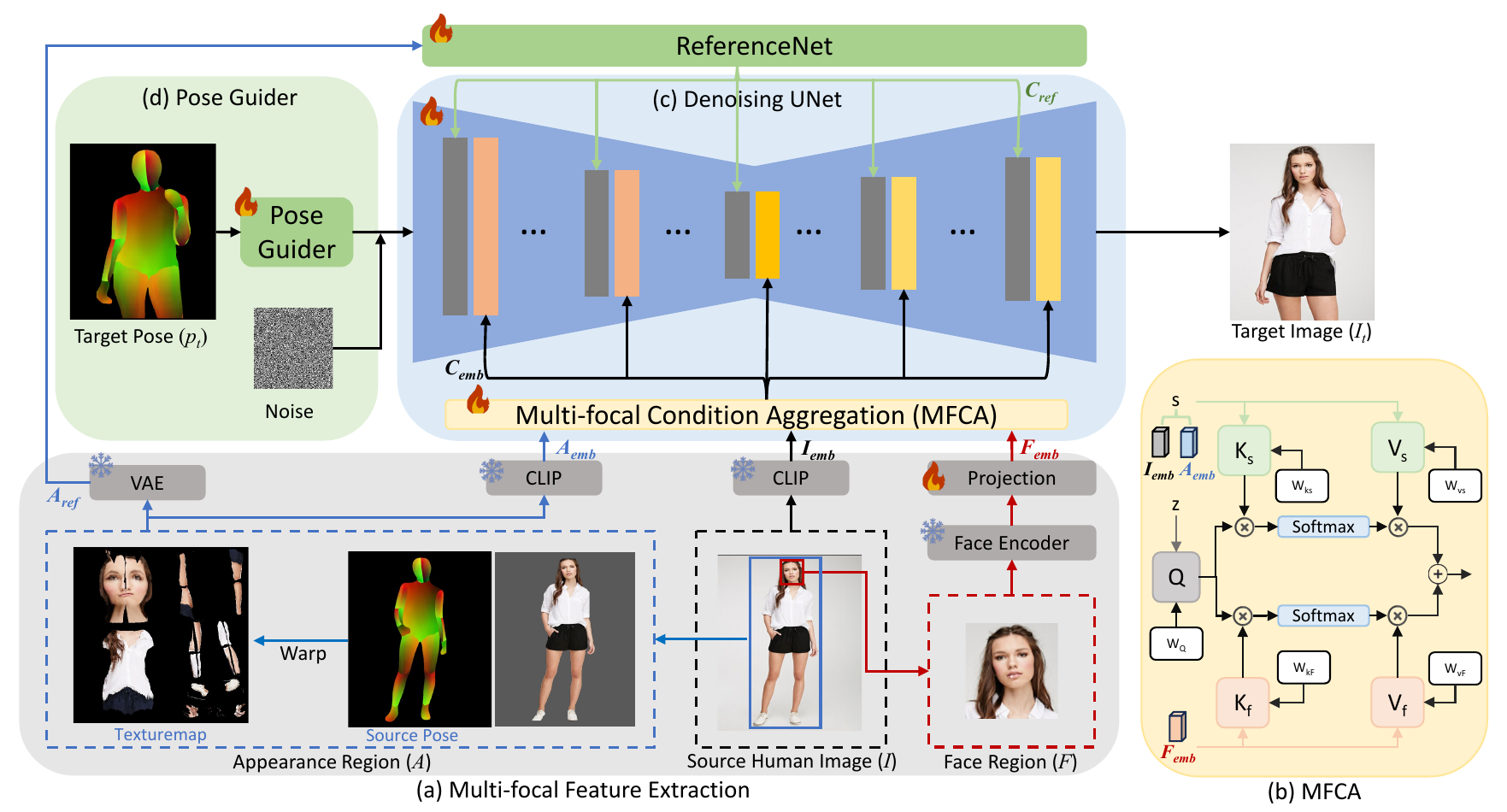}
\caption{The overall pipeline of our proposed Multi-focal Conditioned Diffusion Model. 
(a) Face regions and appearance regions are first extracted from the source person images; (b) multi-focal condition aggregation module $\phi$ is used to fuse the focal embeddings as $c_{emb}$; (c) ReferenceNet $\mathcal{R}$ is used to aggregate information from the appearance texture map, denoted as $c_{ref}$; (d) Densepose provides the pose control to be fused into UNet with noise by Pose Guider.}

\label{fig::workflow}
\vspace{-2mm}
\end{figure*}

\noindent\textbf{Conditional Diffusion Models.}
Recently, diffusion models~\cite{ho2020denoising, pmlr-v202-song23k} have outperformed GANs and significantly improved the visual fidelity of synthesized images across various domains, including text-to-image generation~\cite{rombach2022high,saharia2022photorealistic,ruiz2023dreambooth}, person image generation~\cite{bhunia2023person,lu2024coarse,han2023controllable,xue2024self,cheong2023upgpt}, and 3D avatar generation~\cite{kolotouros2023dreamhuman,huang2023humannorm,liao2023tada,poole2022dreamfusion,liu2023zero1to3}. For most tasks, the widely used model is Stable Diffusion~\cite{rombach2022high} (and its variants), which is a unified conditional diffusion model that allows for semantic maps, text, or images to be used as conditions for controlling generation. Its key contributions lie in applying the diffusion process in latent space, which minimizes resource consumption while maintaining generation quality and flexibility. In this paper, our architecture, along with the baseline’s, is derived from this conditional model, i.e, Stable Diffusion. Previous conditional diffusion models can be categorized into three types based on the condition: text-conditioned~\cite{rombach2022high,ruiz2023dreambooth}, image-conditioned~\cite{hu2024animate,lu2024coarse,kim2024stableviton,mou2024t2i}, and mixed-conditioned models~\cite{ye2023ip,zhang2023adding}. These methods typically use a pretrained model~\cite{rombach2022high, radford2021learning, oquab2023dinov2} to extract condition features, which are then injected into the denoising UNet via cross-attention. Different from the main stream approaches that regard images and texts as a whole, our proposed Multi-focal Conditioned method takes a human image as the input, transforms it by different focuses(e.g., texture maps and facial features), and encodes these focuses into embeddings using various pre-trained models. This approach is a sophisticated combination of image-conditioned and mixed-conditioned strategies. Additionally, we introduce a Multi-focal Conditions Aggregation technique to effectively distribute these conditions into the UNet.

\section{Methodology}
\label{sec:method}


Given a reference image $\mathcal{I}$ representing the appearance condition $c$, the task of PGPIS aims to generate a target image $\mathcal{I}_{t}$ with a desired pose $p_{t}$. This is achieved by learning a conditional network $\mathcal{T}$ such that $\mathcal{I}_{t} = \mathcal{T}(c, p_{t})$. 
While the representation of $p_{t}$ is typically predefined~\cite{guler2018densepose, cao2017realtime, SMPL}, the success of generation largely relies on the network $\mathcal{T}$ and conditions $c$, which extract the shared pose-irrelevant appearance features between $\mathcal{I}$ and $\mathcal{I}_{t}$, ensuring high-quality synthesis of $\mathcal{I}_t$.
To enhance synthesis, we introduce a diffusion model $\epsilon_{\theta}$ conditioned by multiple factors, collectively denoted as $c^*$, which iteratively recovers $\mathcal{I}_t$ from noise.


\subsection{Multi-Conditioned Latent Diffusion Model}

The backbone of our proposed method is based on Stable Diffusion \cite{rombach2022high} (SD), which is an implementation of LDM. 
An encoder $\mathcal{E}$ compresses the image $\mathcal{I}$ to a latent $z$,  and a decoder $\mathcal{D}$ transforms $z$ back to an image $\mathcal{I}' = \mathcal{D}(z)$. The compressed latent representation reduces the optimization spaces and allows the generation of higher resolution and richer diversity. The optimization of loss $\mathcal{L}$ in LDM can be repurposed as: 

\begin{equation} 
\mathcal{L}_{mse} = \mathbb{E}_{z_{t}, p, t, \epsilon, c^{*}}(|| \epsilon - \epsilon_{\theta}(z_{t}, p_t, t, c^{*})||),
\end{equation}
where $\epsilon_{\theta}$ represents the forward process of UNet in LDM, $p_{t}$ is the target pose, $z_t$ is the noisy latent $z$ under timestep $t$, and $c^*$ is our multi-focal condition. 

Despite the advantages of having a latent representation, $\mathcal{I}'$ deteriorates during the compression process. While the perceptual differences between $\mathcal{I}'$ and $\mathcal{I}$ may be very small, this degeneration diminishes the significance of the latent code $z$, particularly for features that are supposed to exhibit substantial variance in the original input $\mathcal{I}$, such as facial traits and garment texture. Furthermore, this deterioration is further magnified since $\mathcal{L}_{mse}$ of $\mathcal{T}$ could not guarantee the generated latent without any deviation, and finally results in an unsatisfactory appearance generation results in these high-frequency regions. 
Previous LDM-based methods~\cite{lu2024coarse, han2023controllable} have neglected this issue by relying only on images, which resulted in the model's failure to accurately generate sensitive regions.

To address this problem, we propose a solution that utilizes multi-focal conditions $c^*$ to focus attention on the important areas of the image. To implement this approach, we have designed a two-branch conditional diffusion model that effectively captures multi-focal attention.
The pipeline is shown in Fig.\ref{fig::workflow}. On the first branch, we follow the structure of ReferenceNet~\cite{hu2024animate} to provide the semantic and low-level features $c_{ref}$, which are concatenated with the UNet features in each stage. In the second branch, we exploit pretrained models to embed three selected focal features from the source image $\mathcal{I}$, face region $\mathcal{F}$, and appearance region $\mathcal{A}$, respectively. These embeddings are aggregated into UNet with our Multi-focal Conditions Aggregation (MFCA).


\subsection{Multi-focal Conditions Aggregation.}

\noindent \textbf{Multi-focal Regions.} To enhance latent feature preservation, we incorporate high-frequency focal regions $c^*$ from the source image $\mathcal{I}$ as conditioning inputs. These focal regions help guide attention mechanisms to mitigate the degradation of human-sensitive features. In our implementation, we focus on regions of the face and appearance that, although they constitute a small portion of the image, capture essential perceptual variations.  The degradation of these areas within the autoencoder can lead to losing fine details, potentially causing the latent feature representation to overlook subtle distinctions present in the source image.

Specifically, we employ~\cite{liu2016ssd} to crop the source image $\mathcal{I}$ obtaining the face region $F$. Additionally, we attain the appearance region $\mathcal{A}$ by warping $\mathcal{I}$ into a structured texture map defined by the SMPL model~\cite{SMPL}, indexing from its estimated DensePose~\cite{guler2018densepose} $p_{\mathcal{I}}$. The texture map disentangles appearance from the posed image, retaining only the pose-invariant texture information.

\noindent \textbf{Multi-focal Embeddings.} 
The three multi-focal conditions are managed using pretrained modules. Starting with a source image $\mathcal{I}$, we extract its embedding $\mathcal{I}_{emb}$ using a pretrained CLIP image encoder~\cite{radford2021learning}. The texture map $\mathcal{A}$ is processed in two ways through $\mathcal{T}$. First, we encode $\mathcal{A}$ with a VAE encoder~\cite{rombach2022high}, producing an output $\mathcal{A}_{ref}$, which is then passed to ReferenceNet $\mathcal{R}$. Additionally, we use CLIP to obtain the texture map encoding $\mathcal{A}_{emb}$. For facial regions $\mathcal{F}$, we note that general image encoders like CLIP may struggle to accurately capture identity features, as face appearance and view in $\mathcal{I}_t$ may differ significantly from those in $\mathcal{I}$.
To address this, we utilize a pretrained face recognition model~\cite{deng2019arcface} to localize the face region and extract identity features. These features are then projected to match the dimensionality of the previous embeddings, noted as $\mathcal{F}_{emb}$. It’s important to note that both $\mathcal{F}_{emb}$ and $\mathcal{A}_{emb}$ are shared between $\mathcal{I}$ and $\mathcal{I}_t$, as they are pose-invariant and represent attributes of the same appearance.

%

\noindent \textbf{Multi-focal Conditioning.}
The conditions $c^*$ are assembled as follows:
\begin{equation} 
    c^* = \left\{\begin{array}{l}
    c_{ref} = \mathcal{R}(\mathcal{A}_{ref}) \\ 
    c_{emb} = \phi(\mathcal{I}_{emb}, \mathcal{A}_{emb}, \mathcal{F}_{emb}, z), \\ 
    \end{array}\right.
\end{equation}
where $\mathcal{R}$ is a trainable ReferenceNet extracting both the structured details and layouts of appearance regions. $\phi$ denotes a mulit-focal condition aggregation module (MFCA) to aggregate the embeddings to UNet. $z$ is a latent input in UNet. Inspired by InstantID~\cite{wang2024instantid}, $\phi$ is defined as follows (see Fig.\ref{fig::workflow}(b)):



\begin{equation}
\begin{aligned}
\phi = & \sum_{i \in \{s, F_{emb}\}} \lambda_{i} Attn(Q, K_i, V_i), \\
          \quad Q= & zW_q, K_i=iW_{ki}, V_i=iW_{vi}, \\
\end{aligned}
\end{equation}
where $Q$, $K_i$, $V_i$ are query, key and value matrices for cross-attentions. $W$ is the attention weight and $\lambda_{i}$ is the scaling factor. $Q$ is computed from latent $z$ while $K_i$, $V_i$ are computed from conditioning embeddings $i$, including the face $F_{emb}$ and a selective condition switcher $s$. $s$ is defined as:
\begin{equation}
    s = \left\{\begin{array}{ccc}
    \mathcal{I}_{emb} & if & z \in \mathcal{U}_{\mathcal{E}} \\
    cat(\mathcal{I}_{emb}, \mathcal{A}_{emb}) & if & z \in \mathcal{U}_{\mathcal{M}} \\
    \mathcal{A}_{emb} & if & z \in \mathcal{U}_{\mathcal{D}} \\
    \end{array}\right.
\end{equation}
where $\mathcal{U}_{\mathcal{E}}$, $\mathcal{U}_{\mathcal{M}}$, $\mathcal{U}_{\mathcal{D}}$ are the encoder, the latent layer and decoder of UNet, respectively.

When combining all conditions, our Multi-Focal Condition Aggregator (MFCA) efficiently aggregates the multi-focal embeddings. This efficiency stems from reducing attention operations to focus on a specific region at each step, while simultaneously leveraging the embedding properties and the inherent structure of the UNet architecture.

Moreover, we introduce a selective condition injection approach to accommodate the distinct characteristics of the UNet structure. Specifically, $\mathcal{U}_{\mathcal{E}}$ encodes information into a lower-dimensional space, where injecting global information from $\mathcal{I}_{emb}$ related to high-level semantics, such as cloth categories, and general background. Conversely, during the decoding stage, $\mathcal{U}_{\mathcal{D}}$ requires fine-grained information to effectively reconstruct the final image; thus, $\mathcal{A}_{emb}$ are injected to provide pose-irrelevant features, such as texture details and garments details, at this phase to fulfill that need. This targeted injection strategy reduces parameter counts and guides the model to prioritize the information most relevant to each specific architectural stage.



Since $\mathcal{F}_{emb}$ is derived from a pretrained face recognition model, it maintains robustness across diverse views and appearances. We retain $\mathcal{F}_{emb}$ throughout all stages of UNet to consistently represent both the input and target faces. An addition operation is employed to aggregate $\mathcal{F}_{emb}$ and $s$.

\noindent \textbf{Pose Guider.} We harness Densepose as our pose condition as it provides appropriate 3D information as claimed in PoCoLD~\cite{han2023controllable}. In addition, Densepose coordinates establish a bijection between the UV space texture map $\mathcal{A}$ and image pixels of $\mathcal{I}_t$, which implicitly bridges the appearance alignment for the two focuses. Similar to \cite{hu2024animate}, we employ a lightweight pose guider module constructed with a series of convolutional layers derived from ControlNet. This module is initialized with pretrained parameters from the ControlNet segmentation model, enabling it to leverage prior knowledge for enhanced guidance. 

\vspace{-3mm}
\subsection{Overall objective}
To force the model to concentrate more on the target face regions, we introduce an extra loss for supervision at face regions:
\begin{equation} 
\mathcal{L}_{face} = \mathbb{E}_{z_{t}, p, t, \epsilon, c^{*}}(|| (\epsilon - \epsilon_{\theta}(z_{t}, p_t, t, c^{*})) \odot m ||)
\end{equation}
where $m$ is the segmentation mask of face regions, which is parsed from the densepose $p_t$. 

Combining eq.(1), the overall objective function is:
\begin{equation} 
\mathcal{L}_{overall} = \mathcal{L}_{mse} + \mathcal{L}_{face} 
\end{equation}
\begin{figure*}[!t]
\centering
\includegraphics[width=1\linewidth]{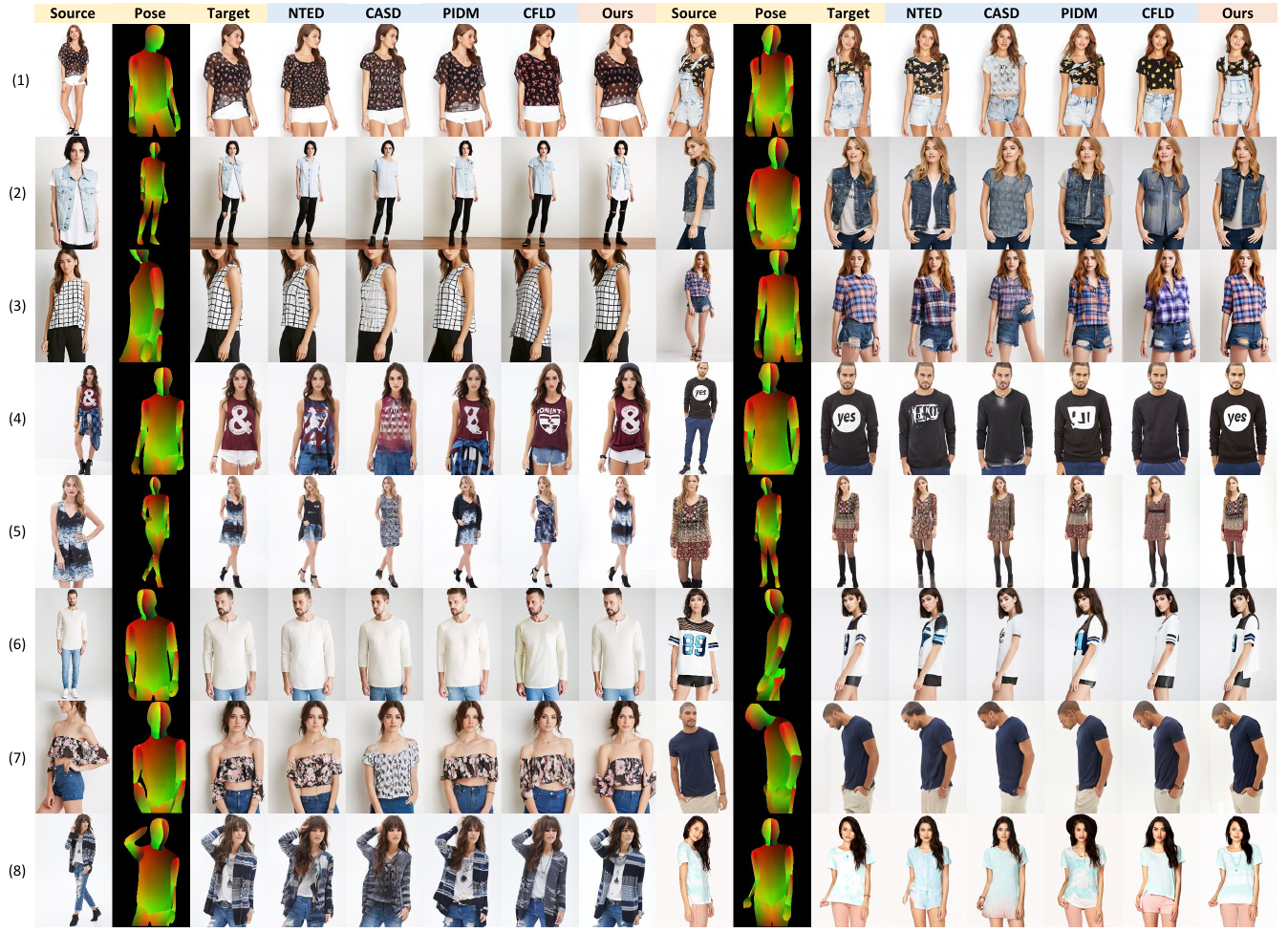}
\caption{Qualitative Comparison with several state-of-the-art models on the Deepfashion dataset. The inputs to our models are the target pose $p_{t}$ and the source person image $\mathcal{I}$. From left to right the results are of NTED, CASD, PIDM, CFLD and ours respectively.}
\label{fig::qualitative}
\end{figure*}

\vspace{-4mm}
\section{Experiments}
\label{sec:exp}
In this section, we present a detailed analysis of our experiments including the dataset setup, evaluation metrics, implementation details, and a thorough comparison of our approach with state-of-the-art methods.

\begin{table}[t]
\centering
\resizebox{\linewidth}{!}
{
\begin{tabular}{c|c|c|c|c}
\hline
 Methods & FID$\downarrow$  & LPIPS$\downarrow$ & SSIM $\uparrow$ & PSNR$\uparrow$\\
\hline
\rowcolor{lightgray}
\multicolumn{5}{l}{\textit{Evaluation on 256 $\times$ 176 resolution}} \\
GFLA$\ddag$~\cite{ren2020deep} (CVPR20) & 9.827 & 0.1878 & 0.7082 & -- \\
SPGNet$\ddag$~\cite{lv2021learning} (CVPR21) & 16.184 & 0.2256 & 0.6965 & 17.222 \\
NTED$\ddag$~\cite{ren2022neural} (CVPR22) & 8.517 & 0.177 & 0.7156 & 17.74 \\
CASD$\ddag$~\cite{zhou2022cross}(ECCV22) & 13.137 & 0.1781 & 0.7224 & 17.880 \\
PIDM$\dag$~\cite{bhunia2023person} (CVPR23) & \textbf{6.36} & 0.1678 & 0.7312 & -- \\
PoCoLD$\dag$~\cite{han2023controllable} (ICCV23) & 8.067 & 0.1642 & 0.7310 & -- \\
CFLD$\dag$ (CVPR24) & 6.804 & \underline{0.1519} & \underline{0.7378} & \underline{18.235} \\
\hline
MCLD (B3)  & 6.86 & 0.157 & 0.734 & 18.03 \\
MCLD (Ours)  & \underline{6.693} & \textbf{0.1482} & \textbf{0.7511} & \textbf{18.84} \\
\hline
\rowcolor{lightgray}
\multicolumn{5}{l}{\textit{Evaluation on 512 $\times$ 352 resolution}} \\
CoCosNets~\cite{zhou2021cocosnet} (CVPR22) & 13.325 & 0.2265 & 0.7236 & -- \\
NTED$\ddag$~\cite{ren2022neural} (CVPR22) & 7.645 & 0.1999 & 0.7359 & 17.385 \\
PIDM$\dag$~\cite{bhunia2023person} (CVPR23) & \textbf{5.8365} & \underline{0.1768} & 0.7419 & -- \\
PoCoLD$\dag$~\cite{han2023controllable} (ICCV23) & 8.416 & 0.1920 & 0.7430 & -- \\
CFLD$\dag$~\cite{lu2024coarse} (CVPR24) & 7.149 & 0.1819 & \underline{0.7478} & \underline{17.645} \\
\hline
MCLD (B3)  & 7.23 & 0.1951 & 0.7405 & 17.48 \\
MCLD (Ours) & \underline{7.079} & \textbf{0.1757} & \textbf{0.7557} & \textbf{18.211} \\
\hline
\end{tabular}
}
\caption{Qualitative comparison with the state-of-the-arts in terms of image quality benchmarks. $\dag$ The scores are reported in their paper, since the same split is followed. $\ddag$ The scores are evaluated and reported in CFLD~\cite{lu2024coarse}, since they split validation set in a different way. Our evaluation code is the same as CFLD.}\label{tab::qualitative}
\vspace{-3mm}
\end{table}

\noindent \textbf{Dataset.} Following~\cite{lu2024coarse, bhunia2023person, han2023controllable, zhu2019progressive}, experiments are conducted using the DeepFashion In-Shop Clothes Retrieval Benchmark~\cite{liu2016deepfashion}, which contains 52,712 high-resolution images of fashion models. Consistent with the CFLD, we split the dataset into training and validation subsets, comprising 101,966 and 8,570 non-overlapping image pairs, respectively. Pose pairs are extracted by Densepose and we evaluate results on 256$\times$176 and 512$\times$352 resolutions.

\noindent \textbf{Metrics.} We conduct two groups of objective metrics to evaluate the overall generated image quality and the generated face preservation, respectively. For the overall generated image quality, four metrics are adopted for comparison. The Fréchet Inception Distance (FID)~\cite{NIPS2017_7240} measures the Wasserstein-2 distance between the feature distributions of generated images and real images, with features extracted from the Inception-v3 pretrained network. Specifically, the generated image features come from the validation dataset, while the real image features are from the training dataset. The Learned Perceptual Image Patch Similarity (LPIPS)~\cite{zhang2018unreasonable} computes image-wise similarity in the perceptual feature space. Both FID and LPIPS assess image quality in a high-level feature domain. Additionally, we use two pixel-wise metrics: the Structural Similarity Index Measure (SSIM) and Peak Signal-to-Noise Ratio (PSNR), which evaluate the accuracy of pixel-wise correspondences between the generated and real images. To assess the identity preservation of the face region, we use a pretrained Face Recognition Model~\cite{deng2019arcface} to extract the face embeddings and compute the face cosine similarity $FS$ and euclidean distance $dist$ between the face regions of generated images and real images. Both the source image $ref$ and the target image $tgt$ are evaluated to assess the overall model ability.

\begin{table}[t]
\centering
\resizebox{\linewidth}{!}
{
\begin{tabular}{c|c|c|c|c}
\hline
 Methods & FS$_{ref}$$\uparrow$ & dist$_{ref}$$\downarrow$ & FS$_{tgt}$$\uparrow$ & dist$_{tgt}$$\downarrow$  \\
\hline
\rowcolor{lightgray}
\multicolumn{5}{l}{\textit{Evaluation on 256 $\times$ 176 resolution}} \\
CASD~\cite{zhou2022cross} (ECCV22) & 0.207 & 28.80 & 0.317 & 26.28  \\
PIDM~\cite{bhunia2023person} (CVPR23) & 0.270 & 28.06 & 0.394 & 25.17 \\
CFLD~\cite{lu2024coarse} (CVPR24) & 0.243 & 28.86 & 0.363 & 26.11 \\
MCLD (B3) & 0.279 & 28.1 & 0.381 & 25.7 \\
MCLD (Ours)  & \textbf{0.301} & \textbf{27.65} & \textbf{0.413} & \textbf{25.02}  \\
$ref$& -- & --  & 0.497 & 22.53 \\
\hline
\rowcolor{lightgray}
\multicolumn{5}{l}{\textit{Evaluation on 512 $\times$ 352 resolution}} \\
CFLD~\cite{lu2024coarse} (CVPR24) & 0.227 & 28.62 & 0.286 & 27.54 \\
MCLD (B3) & 0.289 & 27.04 & 0.333 &  \textbf{26.25} \\
MCLD (Ours) & \textbf{0.294} & \textbf{27.01} & \textbf{0.344} & 26.31 \\
$ref$ & -- & -- & 0.643 & 17.42 \\
\hline
\end{tabular}
}
\caption{Qualitative comparison with the state of the art regarding face quality benchmarks. $FS$ is the face similarity metric, while $dist$ is the euclidean distance measure. $ref$ refer to the input source human image, $tgt$ is the ground truth image. Both $ref$ and $tgt$ are real images.}\label{tab::qualitative_face}
\vspace{-2mm}
\end{table}




\noindent \textbf{Implementation details.} Our model is implemented on Stable Diffusion ~\cite{rombach2022high} 1.5 model using PyTorch~\cite{paszke2019pytorch} and Huggingface Diffusers. The source image and the target image are resized to 512$\times$512. Face regions are detected by a single shot detector~\cite{liu2016ssd} implemented in OpenCV~\cite{itseez2015opencv}, while the face embedding is acquired by a pretrained face analysis model, antelopev2\footnote{https://github.com/deepinsight/insightface}. For appearance regions, the images are first converted to 24 parts defined in Densepose with the size of 200$\times$200, then these parts are transformed to 512$\times$512 SMPL texture map by a predefined mapping. The model is trained for 60,000 iterations using Adam optimizer~\cite{kingma2014adam} with a learning rate of 1e-5. We train our model on two Nvidia A100 GPUs with a batch size of 12 for each GPU. During sampling, a classifier-free guidance (CFG) strategy is adopted to improve the sampling quality. We set the CFG scale to 3.5 and $\lambda_i$ in MFCA to 1 and 0.5.

\subsection{Quantitative Comparison}
Our method is compared with both GAN-based and diffusion-based state-of-the-art approaches. Specifically, PIDM~\cite{bhunia2023person} is diffusion based while PoCoLD~\cite{han2023controllable} and CFLD~\cite{lu2024coarse} is LDM-based. The evaluation is performed on two resolutions,  256$\times$176 and 512$\times$352. In addition, we compare our method with our baseline $B3$ since it has an aggregation structure similar to~\cite{wang2024instantid}. As shown in Tab~\ref{tab::qualitative},  our method performs better by conditioning with multi-focal regions in the image quality benchmarks. 
LDM-based methods are known to encounter challenges due to autoencoder compression, which often results in suboptimal FID scores compared to fully diffusion-based approaches. Our proposed method mitigates these limitations, achieving improved FID scores among LDM-based techniques. While certain recent diffusion-based methods do not publicly release their best-performing checkpoints, we report results as stated in their respective publications. Additionally, as demonstrated in Tab.~\ref{tab::qualitative_face}, our method exhibits robust identity preservation across evaluated metrics. The table also includes similarity metrics between the reference source image and target ground truth, where our method achieves performance on par with reference images, which serve as authentic representations providing facial cues to the network.
\subsection{Qualitative Comparison}

We present our comprehensive visual comparison with recent approaches that release their validation results or reproducible, from the left to right is NETD~\cite{ren2022neural}, CASD~\cite{zhou2022cross}, PIDM~\cite{bhunia2023person}, CFLD~\cite{lu2024coarse} and ours respectively. We observe several conclusions listed below. Firstly, current methods are suffering from reconstructing the details of the textures since they only use the source image as condition. This is especially noticed in GAN based methods and LDM based method. This is partially because of the limited details representation ability of GAN, and the information deterioration in LDM. However, after introducing the appearance regions by texture map, our method shows a consistent generation results when the provided information from appearance region and face region is adequate. In rows 1-2, our method preserves better clothing styles even when these styles is rare to be seen in dataset. In rows 3-4, our method also shows a consistent ability to reconstruct the appearance patterns under the given reference images. While other methods are struggled to the details of original patterns. In row 5, for these input images with complex patterns, all the methods fail to reproduce the same details. However, our methods shows a consistent layout of cloths. In addition, identity preservation is one of the most challenging task for current methods, since it is highly sensitive from human perception but not for generative losses. As the illustrated image shows, especially in rows 6-8, our method performs a good identity preserving by introducing the invariant face region embeddings as conditions and supervisions. 

\begin{table}[t]
\centering
\resizebox{\linewidth}{!}
{
\begin{tabular}{c|c|c|c|c|c|c|c}
\hline
 Method & Conditions & Aggregation & Params & FID & LPIPS & SSIM & PSNR\\
\hline
\rowcolor{lightgray}
\multicolumn{8}{l}{\textit{Evaluation on 256 $\times$ 176 resolution}} \\
B1 & $I$ & -- & 1622 M  & \textbf{6.427} & 0.1629 & 0.7371 & 18.18 \\
B2 & $I$,$A$ & concat & 1622 M & 6.830 & 0.1620 & 0.7357 & 18.20 \\
B3 & $I$,$A$,$F$ & concat & 1698 M& 6.858 & 0.1536 & 0.7340 & 18.03 \\
B4 & $F$,$A$ & MFCA & 1711 M& 6.717 & 0.1619 & 0.734 & 18.16 \\
B5 & $I$,$A$ & MFCA & 1622 M  & 6.723 & 0.1483 & 0.7499 & 18.72 \\
Ours  & $I$, $A$, $F$& MFCA & 1717 M  &6.693 & \textbf{0.1482} & \textbf{0.7511} & \textbf{18.84} \\
\hline
\rowcolor{lightgray}
\multicolumn{8}{l}{\textit{Evaluation on 512 $\times$ 352 resolution}} \\
B1 & $I$ & -- & 1622 M & \textbf{6.738} & 0.1923 & 0.7463 &  17.64 \\
B2 & $I$,$A$ & concat & 1622 M & 7.129 & 0.1932 & 0.7433 & 17.63 \\
B3 & $I$,$A$,$F$ & concat &1698 M& 7.23 & 0.1951 & 0.7405 & 17.48 \\
B4 & $F$,$A$ & MFCA & 1711 M & 7.138 & 0.1923 & 0.7408 & 17.56 \\
B5 & $I$,$A$ & MFCA & 1622 M& 7.047 & 0.1766 & 0.7543 & 18.13 \\
Ours & $I$, $A$, $F$& MFCA & 1717 M & 7.079 & \textbf{0.1757} & \textbf{0.7557} & \textbf{18.21} \\
\hline
\end{tabular}
}
\caption{Qualitative comparison for ablation studies. $I$,$A$,$F$ are the embeddings from source images, appearance regions, and face regions respectively. Aggregation column refers to the feature fusion strategy. Params refers to the trainable parameter in network.}\label{tab::ablation}
\vspace{-2mm}
\end{table}

\begin{figure}[!t]
\centering
\includegraphics[width=1\linewidth]{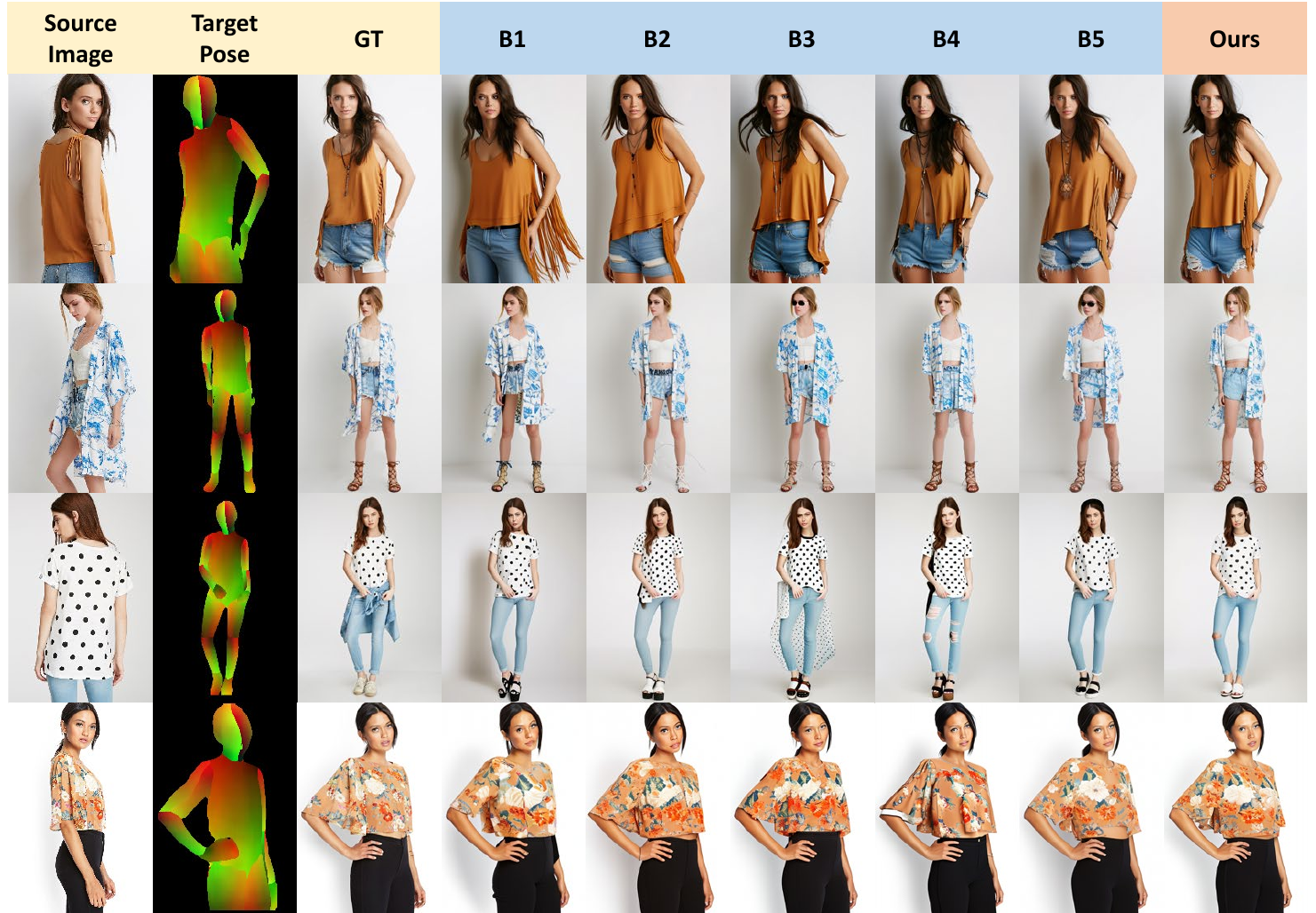}
\caption{Qualitative ablation comparison. Refer to Tab.~\ref{tab::ablation} for baseline settings.}
\label{fig::ablation}
\vspace{-6mm}
\end{figure}

\begin{figure*}[!t]
\centering
\vspace{-3mm}
\includegraphics[width=1\linewidth]{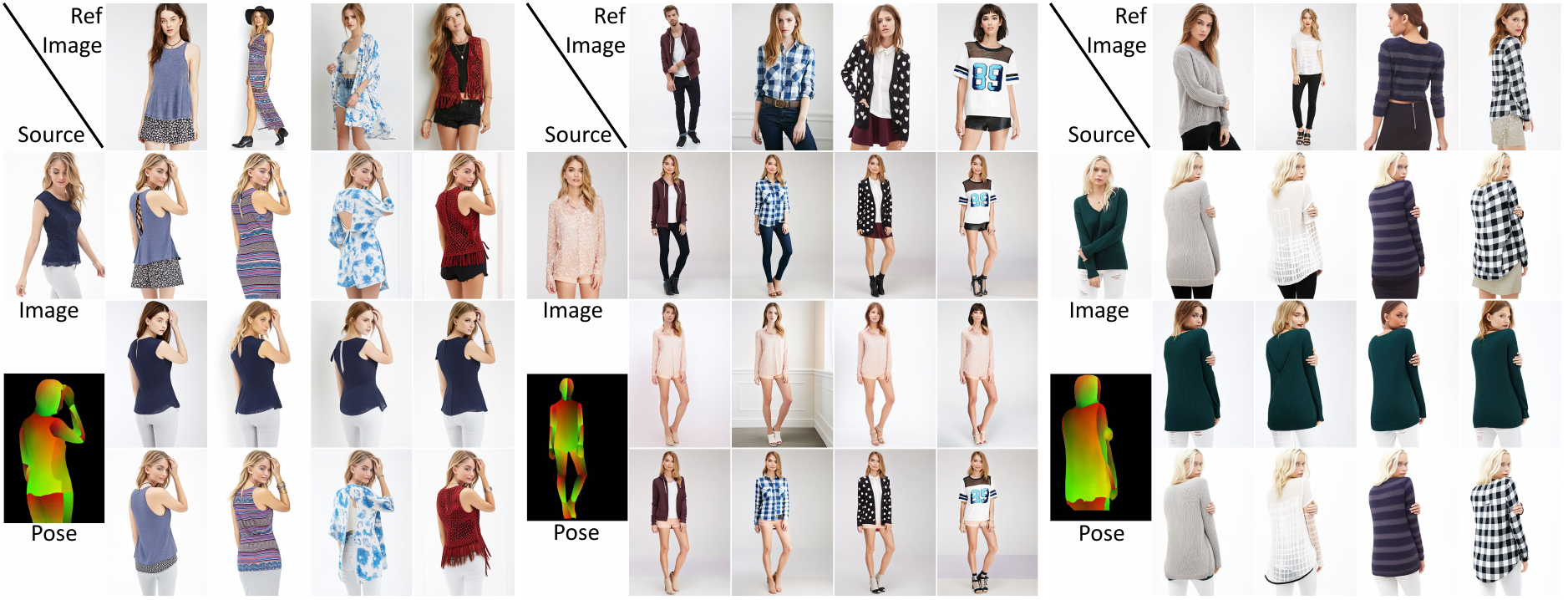}
\caption{ Appearance editing results. Our method accepts flexible editing of given identities, poses, and clothes. This is achieved only by modifying some regions of conditions, and no need for any masking or further training. }
\label{fig::editing}
\vspace{-2mm}
\end{figure*}

\subsection{Ablation Study}

We perform ablation studies on our MFCA module to demonstrate the value of multi-focal conditions. The quantitative result is shown in Tab.~\ref{tab::ablation}, while the qualitative results are illustrated in Fig.~\ref{fig::ablation}. $B1$ only takes $\mathcal{I}_{emb}$ from the source image as a condition, which is similar to other image-based methods. Due to the limited power of the image condition, the generated image fails to preserve facial and textural traits, introducing undesired artifacts. When we gradually add $\mathcal{A}_{emb}$ and $\mathcal{F}_{emb}$ to $B2$ and $B3$ with a concatenation strategy, the cloth style and textures in $B2$ slightly improve. Introducing facial conditioning (\ie $B3$) increasingly improves performance. 
However, this simple concatenation does not ensure stable performance. When too many conditions are handled in parallel, the effort for each condition remains unclear, and the focused areas become ambiguous, resulting in unpredictable cloths styles, textures, and identities. Quantitative results also prove that concatenation struggles to improve the generated image quality. Thus, in $B4$ and $B5$, we adopt our MFCA without the conditions $\mathcal{I}_{emb}$ and $\mathcal{F}_{emb}$, respectively. Overall, this results in a diminishing in unwanted artifacts due to the reduced attention regions.
Qualitatively, when dropping the $\mathcal{I}_{emb}$ in $B4$, the cloth styles lose in detail. $\mathcal{F}_{emb}$ and $\mathcal{A}_{emb}$ only receive the information inside the Densepose estimation, and the regions outside $\mathcal{A}$ are randomly generated which is not consistent to the semantic cloth style. In $B5$ texture improves in quality but the facial traits are almost entirely lost.
This seems to confirm our quantitative findings where the deformed and incomplete warping in the texture map affects the facial appearance. 
Though our method is close to $B5$ in terms of metrics, probably due to the fact that the face regions only occupy a small portion of the entire image, the facial traits are well preserved. 

Finally, we noticed a decrease in FID performance after introducing more conditions. As reported in \cite{lu2024coarse}, the FID of VAE reconstruction in LDM methods is 7.967. Consequently, a lower FID in LDM-based methods does not necessarily indicate a superior overall performance. The other three metrics provide stronger quantitative evidence.

\begin{figure}[!t]
\centering
\includegraphics[width=1\linewidth]{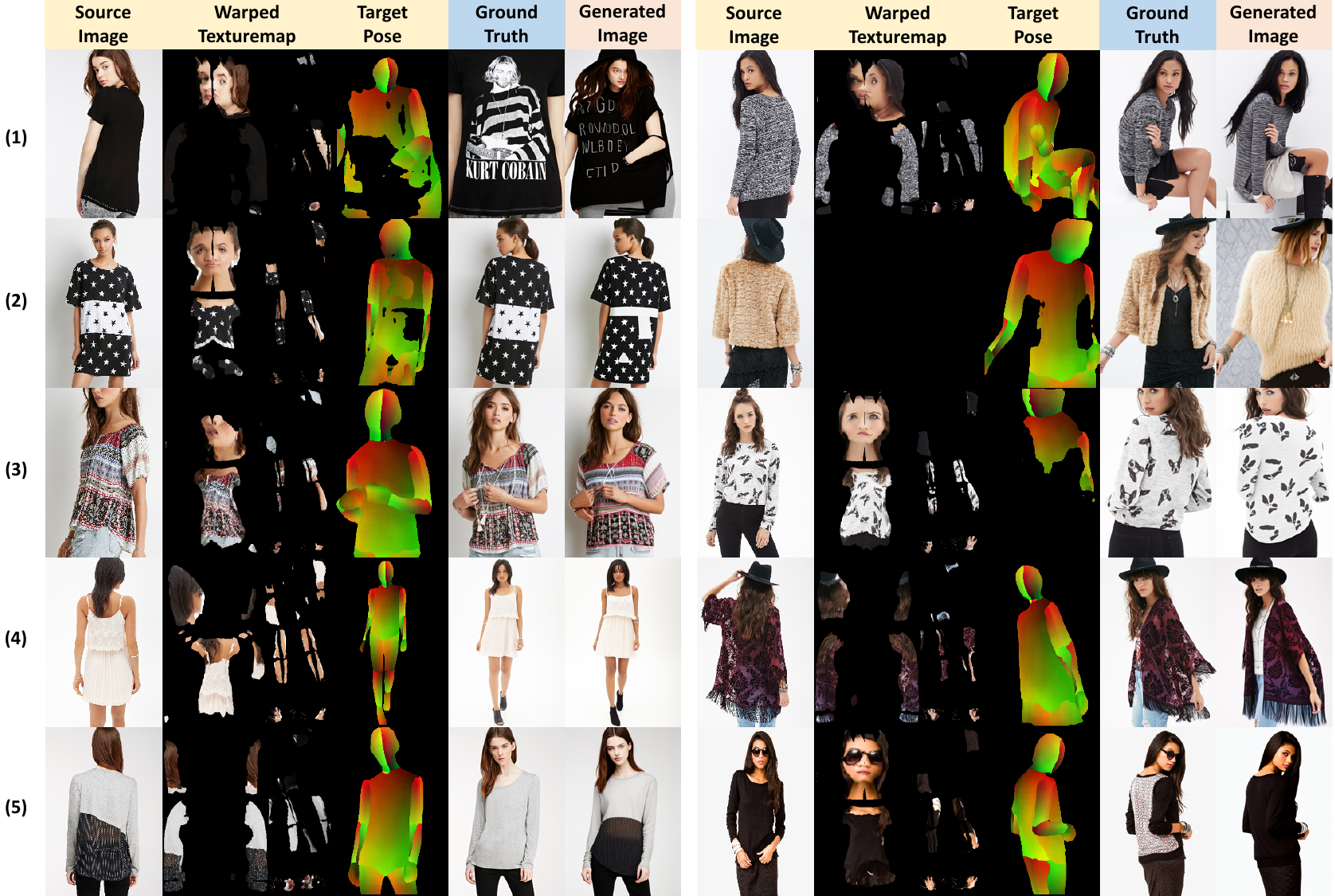}
\caption{Failure cases caused by (1) Wrong target pose, (2) Incomplete texture map, (3) Squeezed texture map, (4) Missing face information, (5) Significant view changes. }
\label{fig::fail}
\vspace{-2mm}
\end{figure}

\subsection{Appearance Editing}

Our approach enables flexible, localized editing by adjusting specific focal conditions within the generative pipeline, allowing precise control over focal regions. Editing examples are illustrated in Fig.~\ref{fig::editing}. 

By modifying the texture map $\mathcal{A}$ for designated clothing regions, we can seamlessly alter clothing to reflect chosen reference styles, showcasing the strong control capability of our texture map focalization (row 2). Additionally, by substituting the face embedding $F_{emb}$ and updating the corresponding facial regions in the texture map, our method supports person identity swapping (row 3). This disentangling of  maps, facial identity, and pose permits arbitrary combinations of identities, clothing styles, and poses. 

For selective edits, such as altering only specific clothing parts, we can replace corresponding regions within the texture map, which is particularly effective for simpler clothing designs with clear texture segments(as shown in the top section of the row 4). Unlike traditional diffusion-based methods, which rely on mask-based blending within latent spaces, our method provides a more streamlined and adaptable editing solution through structured modifications.

In general, our approach offers a more straightforward and flexible editing solution by solely modifying the structured conditions. This highlights the superiority of our proposed Multi-Focal Conditions Aggregation module in terms of editing capabilities. Furthermore, our editing results are more realistic than those of baseline methods~\cite{lu2024coarse, bhunia2023person, han2023controllable}, as they avoid the boundary artifacts often associated with mask-based techniques. A detailed comparison can be found in the supplementary materials.


\subsection{Failure Cases}
Despite achieving satisfactory appearance-preserving ability in most cases, our model occasionally fails to produce desired results when dealing with uncommon or abrupt images, as shown in Fig.~\ref{fig::fail}. We notice several failure scenarios: (1) the target pose is wrongly estimated, (2) the source texture map is missing or incomplete, (3) the source texture map is fully estimated, but its appearance is shifted to limited pixel resolutions. (4) missing facial traits; (5) significant view changes that are not captured by source image.

\section{Conclusions}
In this paper, we introduced the MCLD framework for pose-guided person image generation. We addressed the challenge of compression degradation in LDMs, especially over sensitive regions, by developing a multi-focal conditioning strategy that strengthens control over both identity and appearance. Our MFCA module selectively integrates pose-invariant focal points as conditioning inputs, significantly enhancing the quality of the generated images. Through extensive qualitative and quantitative evaluations, we demonstrate that MFCA surpasses existing methods in preserving both the appearance and identity of the subject. Moreover, our approach enables more flexible image editing through improved condition disentanglement. In future work, we aim to explore 3D priors to further enhance generation consistency and improve appearance fidelity.

\noindent\textbf{Acknowledgement} This work was supported by the MUR PNRR project FAIR (PE00000013) funded by the NextGenerationEU and the EU Horizon project ELIAS (No. 101120237). 
We acknowledge the CINECA award under the ISCRA initiative, for the availability of high-performance computing resources and support.



\small \bibliographystyle{ieeenat_fullname} \bibliography{main}
\clearpage
\setcounter{page}{1}
\maketitlesupplementary

%
\begin{table}[t]
\centering
\resizebox{\linewidth}{!}
{
\begin{tabular}{c|c|c|c|c|c}
\hline
 Method & NTED~\cite{ren2022neural} & CASD~\cite{zhou2022cross} & PIDM~\cite{bhunia2023person}  & CFLD~\cite{lu2024coarse} & Ours \\
 \midrule
Preferences & 16.4$\%$ & 11.0$\%$ & 17.1$\%$ &12.9$\%$ & \textbf{42.6}$\%$\\
\hline
\end{tabular}
}
\vspace{-0.4cm}
\caption{User Study about the preferences of generated images towards ground truths.}\label{tab::user_study}
\vspace{-0.5cm}
\end{table}

\noindent \textbf{User Study.} We conducted a user study to evaluate the image synthesis quality of various methods~\cite{ren2022neural, zhou2022cross, bhunia2023person, lu2024coarse}, focusing on three key aspects: 1) texture quality, 2) texture preservation, and 3) identity preservation. We recruited 45 volunteers, most of whom are Ph.D. students specializing in deep learning and computer vision. Each participant was asked to answer 30 questions, selecting the method that best matched the ground truth based on the defined quality criteria. The results are listed in Tab.~\ref{tab::user_study}. Compared to other methods, our approach achieved the highest preference score of 42.6$\%$, which is 25.5 percentage points higher than the second-best method. This indicates that our method excels in preserving identities and textures based on objective criterion.

\noindent \textbf{Comparison of Editing.} To compare the editing and its flexibility of our method with mask-based method~\cite{lu2024coarse, bhunia2023person, han2023controllable}, we build upon the concept of CFLD~\cite{lu2024coarse} to address the pose-variant appearance editing task, as demonstrated earlier. To enable the mask-based method to modify the corresponding regions, we introduce an additional denoising pipeline to blend the source image under a given pose. Initially, masks for the editing regions are extracted using a human parsing algorithm 
and then integrated into the sampling process. During sampling, the noise prediction, $\tilde{\epsilon}$, is decomposed into two components: $\epsilon^{s}$, predicted by a UNet conditioned on the source image styles, and $\epsilon^{\text{ref}}$, predicted by the same UNet conditioned on the target image styles. Both of the two componets is conditioned by the same give pose. Let $\tilde{\epsilon}_t$ be defined as follows:
\begin{equation}
\tilde{\epsilon}_t = m \cdot \epsilon^{s}_t + (1 - m) \cdot \epsilon^{ref}_{t},
\end{equation}
where $\epsilon^{s}_t$ and $\epsilon^{ref}_{t}$ is the noise at timestep $t$.

\begin{figure*}[!t]
\centering
\includegraphics[width=1\linewidth]{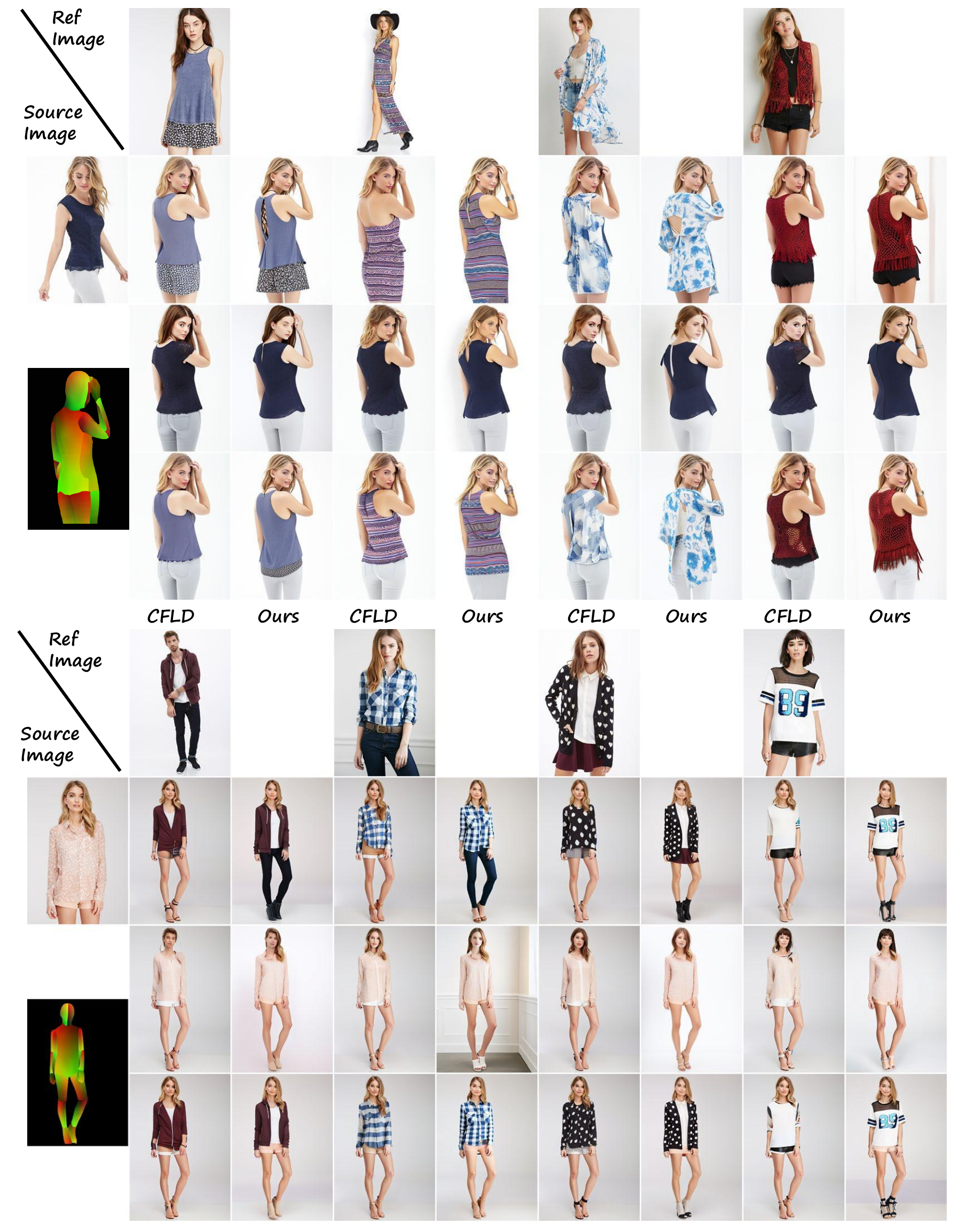}
\caption{Comparison of appearance editing between ours and CFLD. The 2nd, 3rd, 4th rows show the editing of clothes, face, upper cloth region, respectively.}
\label{fig::supp_editing1}
\end{figure*}

\begin{figure*}[t]
\centering
\includegraphics[width=0.95\linewidth]{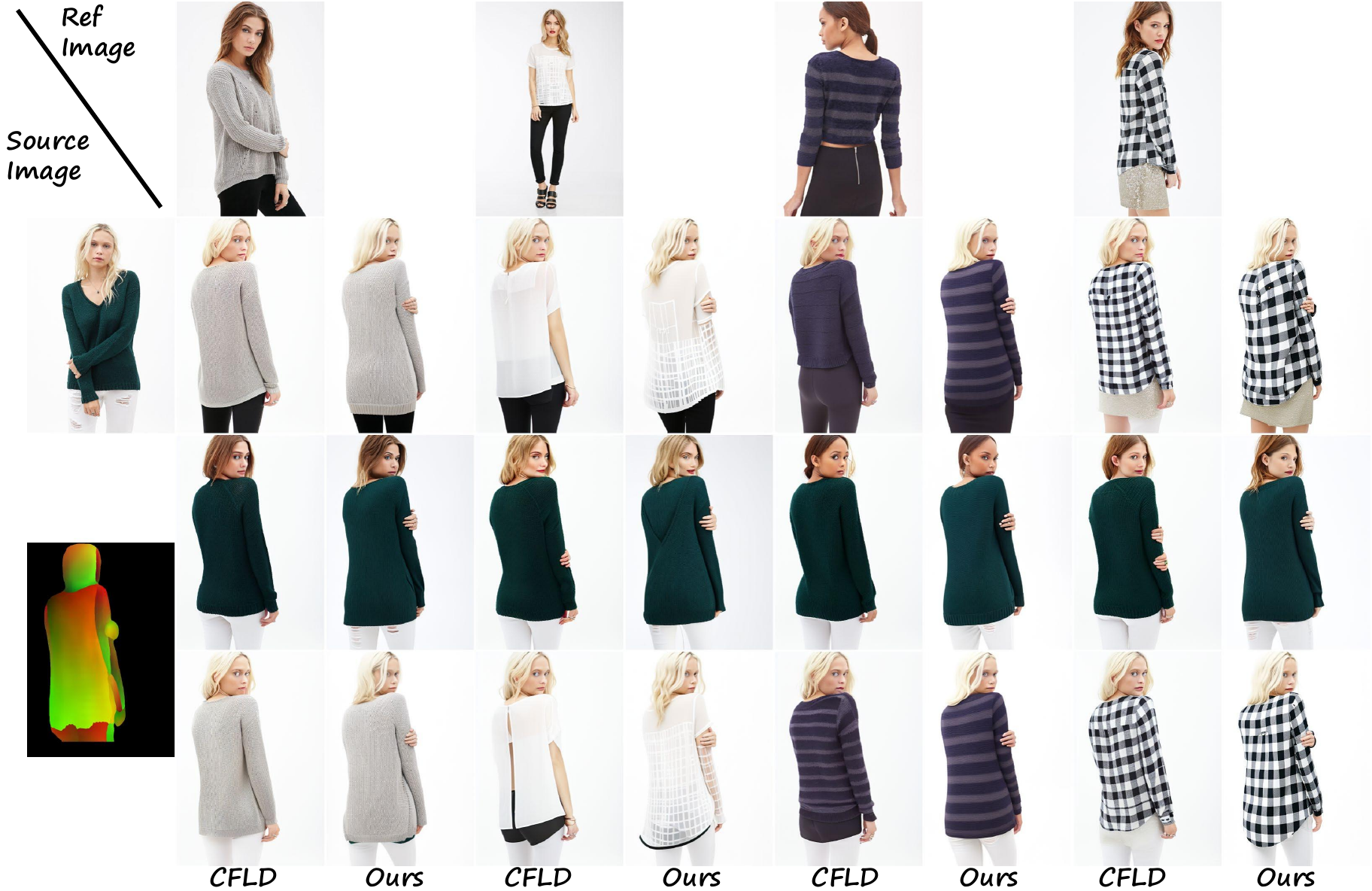}
\vspace{-0.2cm}
\caption{Comparison of appearance editing between ours and CFLD. The 2nd, 3rd, 4th rows show the editing of clothes, face, upper cloth region, respectively.}
\vspace{-0.5cm}
\label{fig::supp_editing2}
\end{figure*}

As shown in Fig.~\ref{fig::supp_editing1},~\ref{fig::supp_editing2}, the method follows the same generation task by separately masking clothes, faces, and upper clothes. However, mask-based methods struggle to preserve facial and texture details under new poses.This limitation arises from the inherent inability of image-conditioned methods to accurately recover fine-grained details. Furthermore, the use of a provided mask introduces additional challenges, as generating precise masks for synthesized images remains non-trivial, often leading to artifacts at the edges in generated images. Moreover, restricting the masked region may adversely affect the preservation of cloth styles and categories, whereas our approach demonstrates superior retention of these attributes.

\begin{figure*}[h]
\centering
\includegraphics[width=0.95\linewidth]{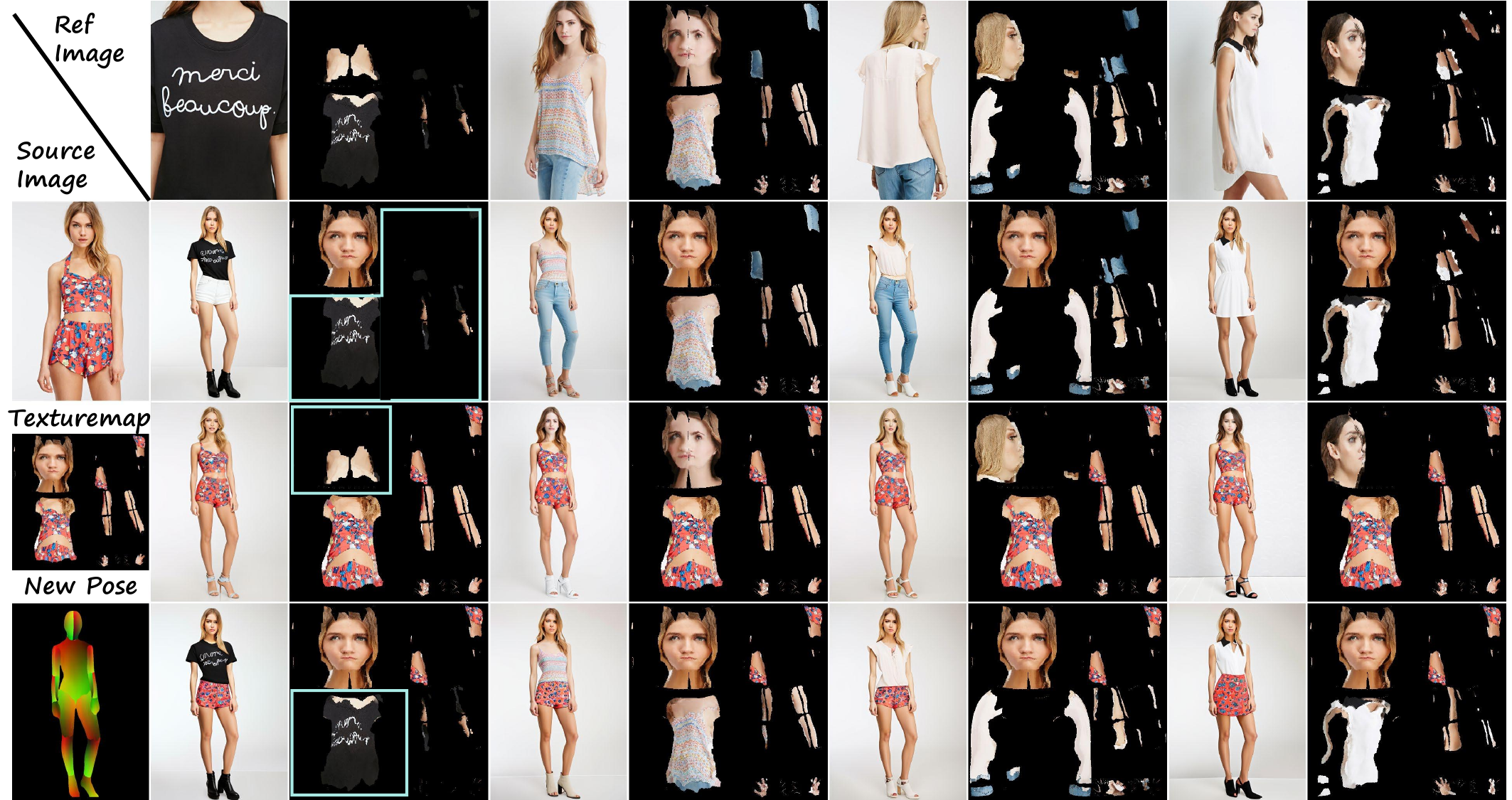}
\vspace{-0.2cm}
\caption{ Additional results of our editings. We show the texture map on the right to illustrate our swapped regions in texture map. The editing regions are labelled using light green bounding box.}
\label{fig::supp8}
\end{figure*}

\begin{figure*}[!t]
\centering
\includegraphics[width=1\linewidth]{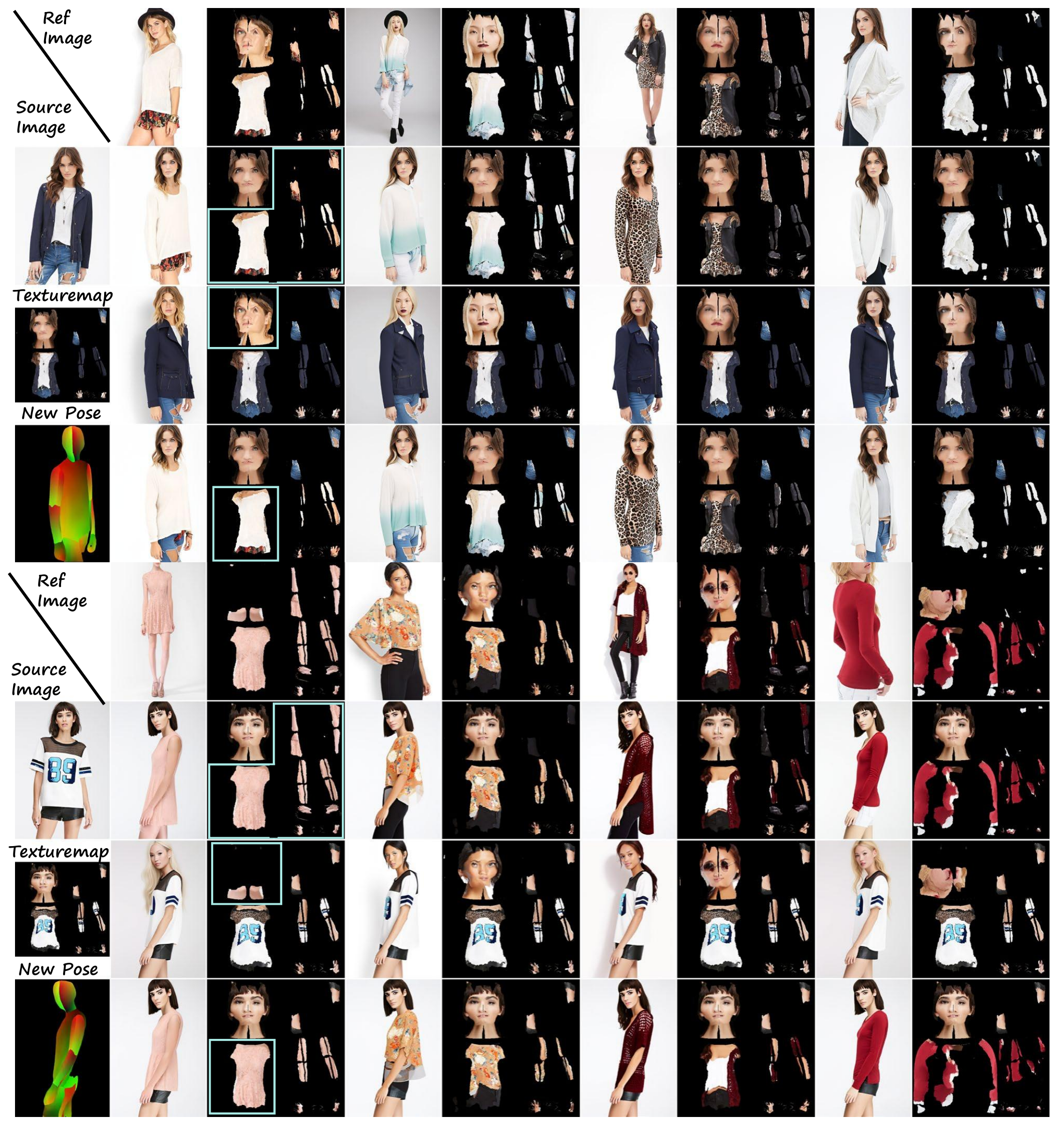}
\caption{ Additional results of our editings. We show the texture map on the right to illustrate our swapped regions in texture map. The editing regions are labelled using light green bounding box.}
\label{fig::supp6}
\end{figure*}

\begin{figure*}[!t]
\centering
\includegraphics[width=1\linewidth]{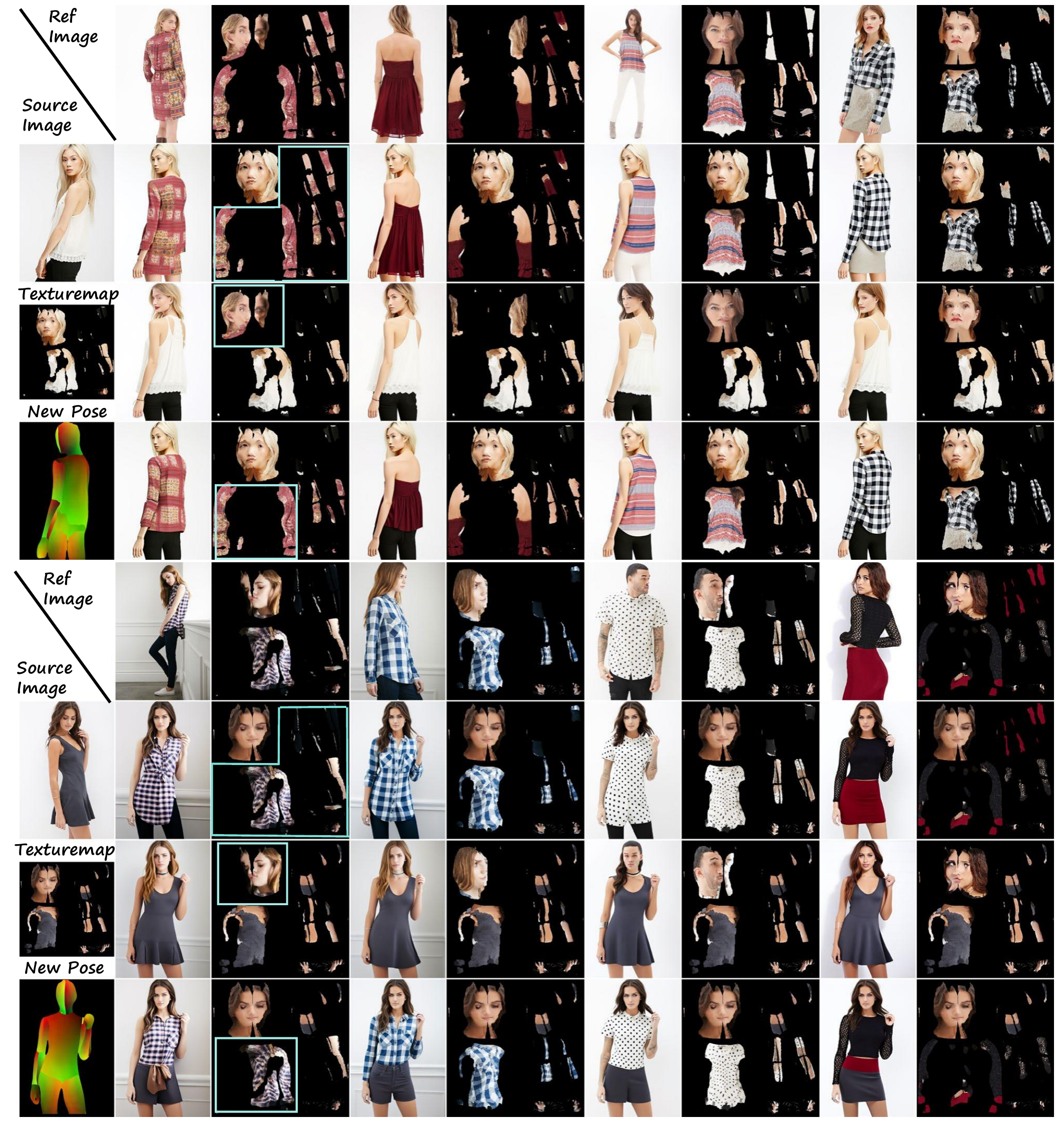}
\caption{ Additional results of our editings. We show the texture map on the right to illustrate our swapped regions in texture map.The editing regions are highlighted with light green bounding boxes.}
\label{fig::supp7}
\end{figure*}

We present additional examples of our generated images in Fig.~\ref{fig::supp8},~\ref{fig::supp6},~\ref{fig::supp7}, illustrating the pose-variant editing setting. For clarity, the swapped texture maps are also provided to highlight the swapping procedures.

In addition, since the edited images feature combined clothing and identities, no ground truth exists in current datasets, making pixel-wise evaluation infeasible. Instead, we provide the quantitative comparison of our method using several perceptual benchmarks in Tab~\ref{tab::editing_metric}, which illustrates our method could generate more natural edited images.

\begin{table}[h]
\centering
\resizebox{\linewidth}{!}
{
\begin{tabular}{c|c|c|c|c}
\hline
Methods & Face Similarity $\uparrow$  &  Face Distance $\downarrow$  &  lpips $\downarrow$  & CLIP score $\uparrow$  \\
\hline
CFLD / Ours & 0.274 / 0.341 & 28.2 / 26.6 & 0.273 / 0.268 &  0.893 / 0.903 \\
\hline
\end{tabular}
}
\vspace{-3mm}
\caption{Perceptual comparison of Editing Performance}\label{tab::editing_metric}
\vspace{-3mm}
\end{table}

\begin{figure}[h]
\vspace{-2mm}
\centering
\includegraphics[width=1\linewidth]{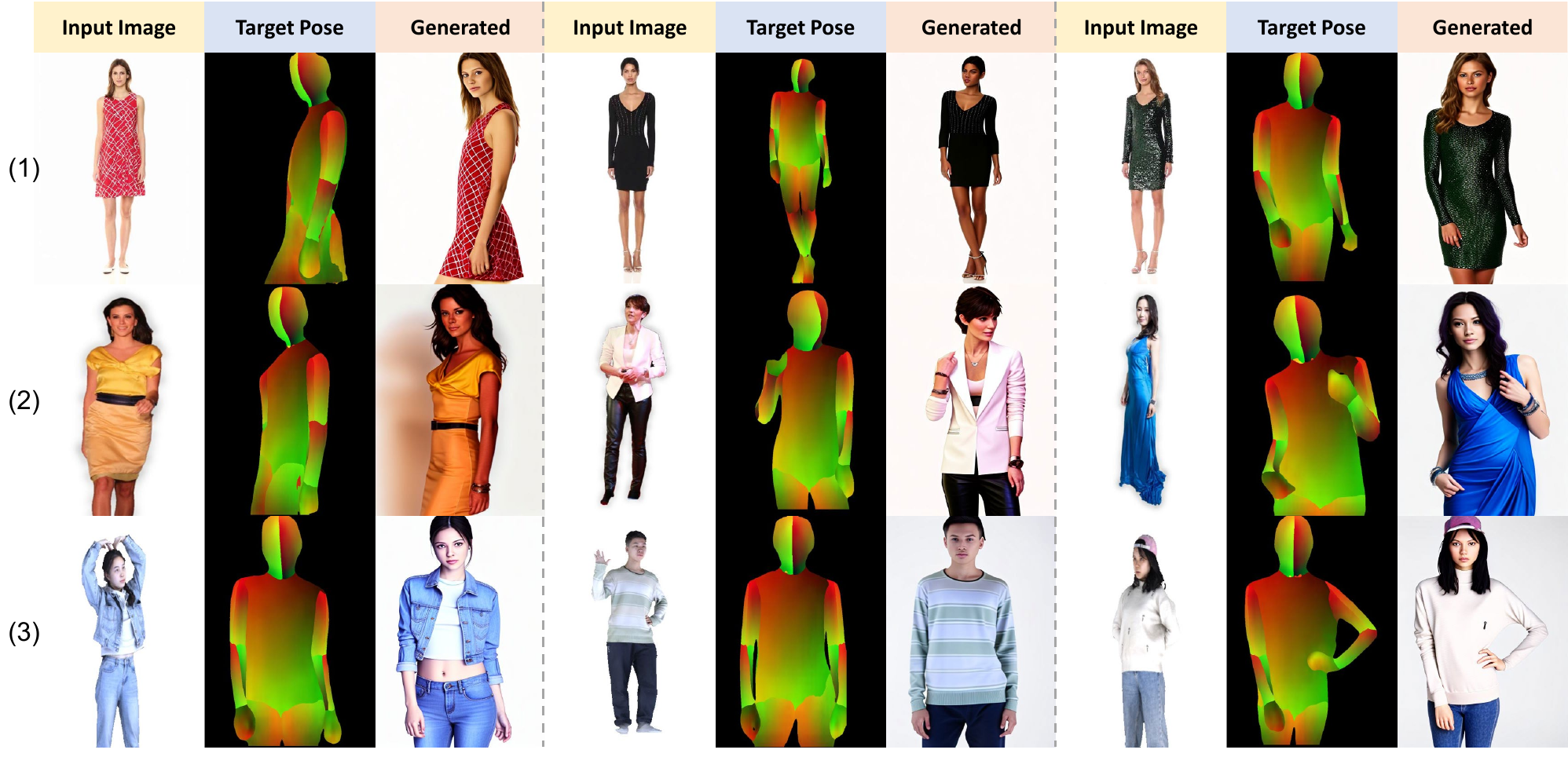}
\label{fig::failure}
\vspace{-5mm}
\caption{Results on other datsets.}\label{fig::general_results}
\vspace{-4mm}
\end{figure}

\noindent \textbf{Generalization to diverse dataset.}  We further validate on 3 out-of-domain datasets without extra model training: 1) UBCFashion~\cite{zablotskaia2019dwnet}, 2) SHHQ~\cite{fu2022stylegan}, 3) Thuman~\cite{yu2021function4d}. We randomly selected some poses and characters from the datasets as input shown in~\ref{fig::general_results}. The results demonstrate consistent, appearance-preserving image generation.

\begin{figure*}[!t]
\centering
\includegraphics[width=1\linewidth]{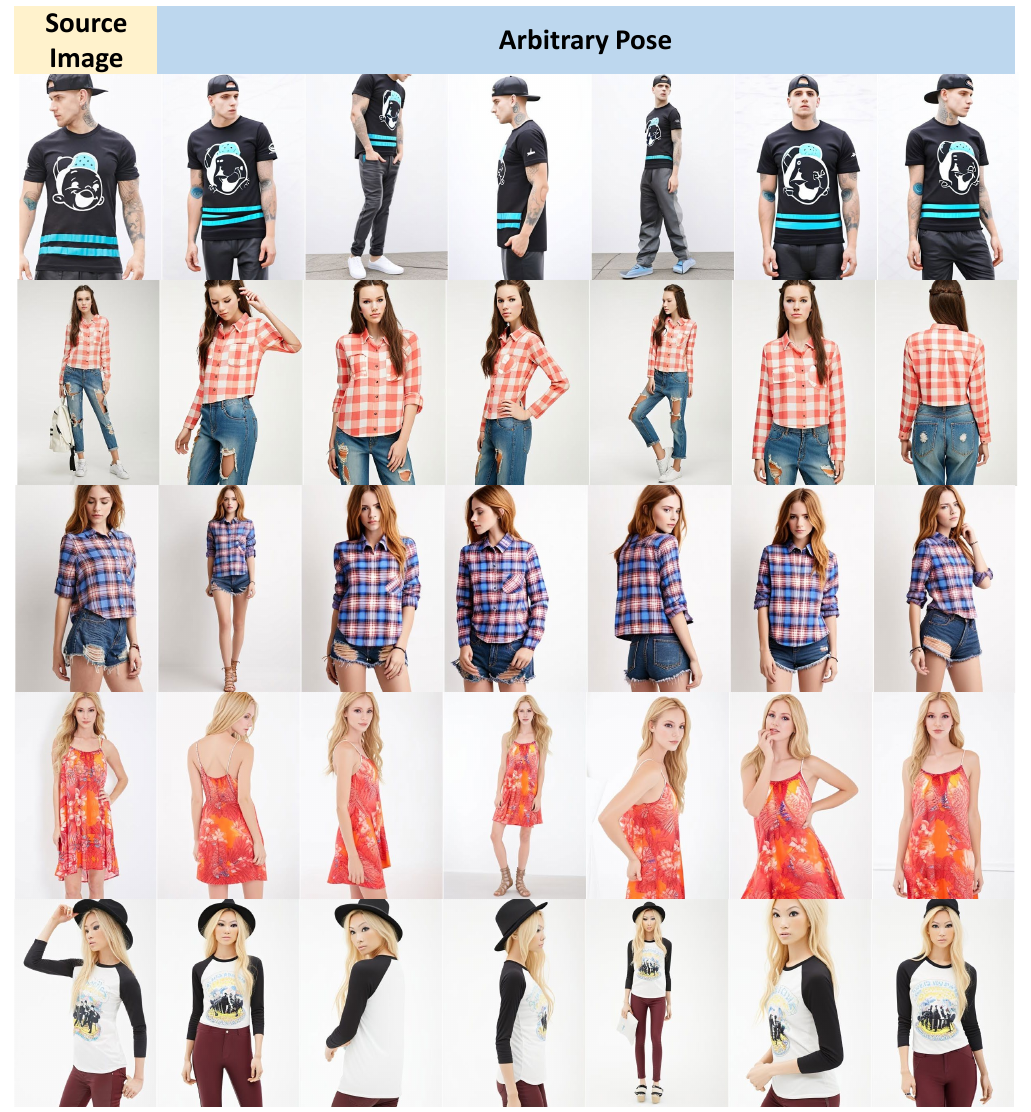}
\caption{ Additional results on arbitrary poses from the test set. }
\label{fig::supp1}
\end{figure*}

\begin{figure*}[!t]
\centering
\includegraphics[width=1\linewidth]{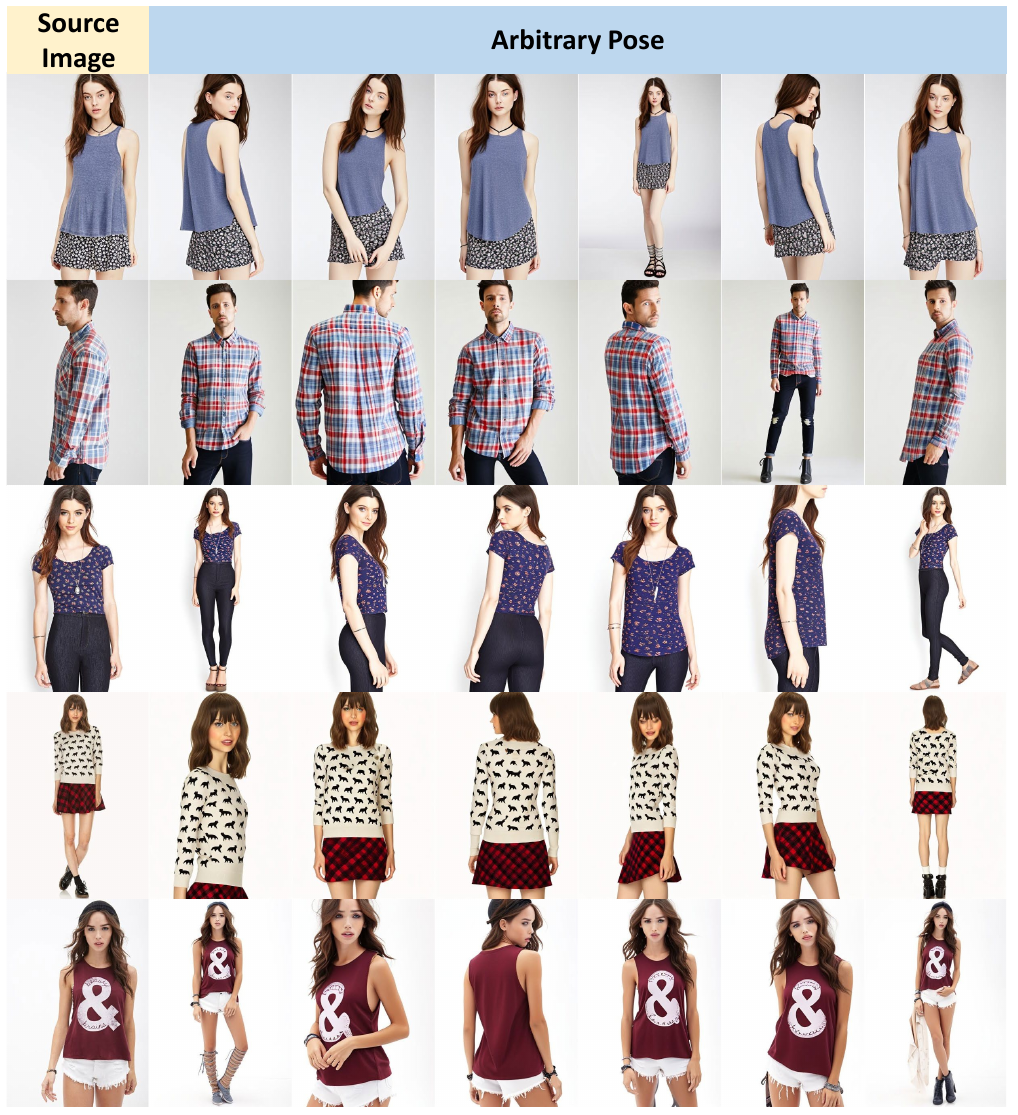}
\caption{ Additional results on arbitrary poses from the test set. }
\label{fig::supp2}
\end{figure*}

\begin{figure*}[!t]
\centering
\includegraphics[width=1\linewidth]{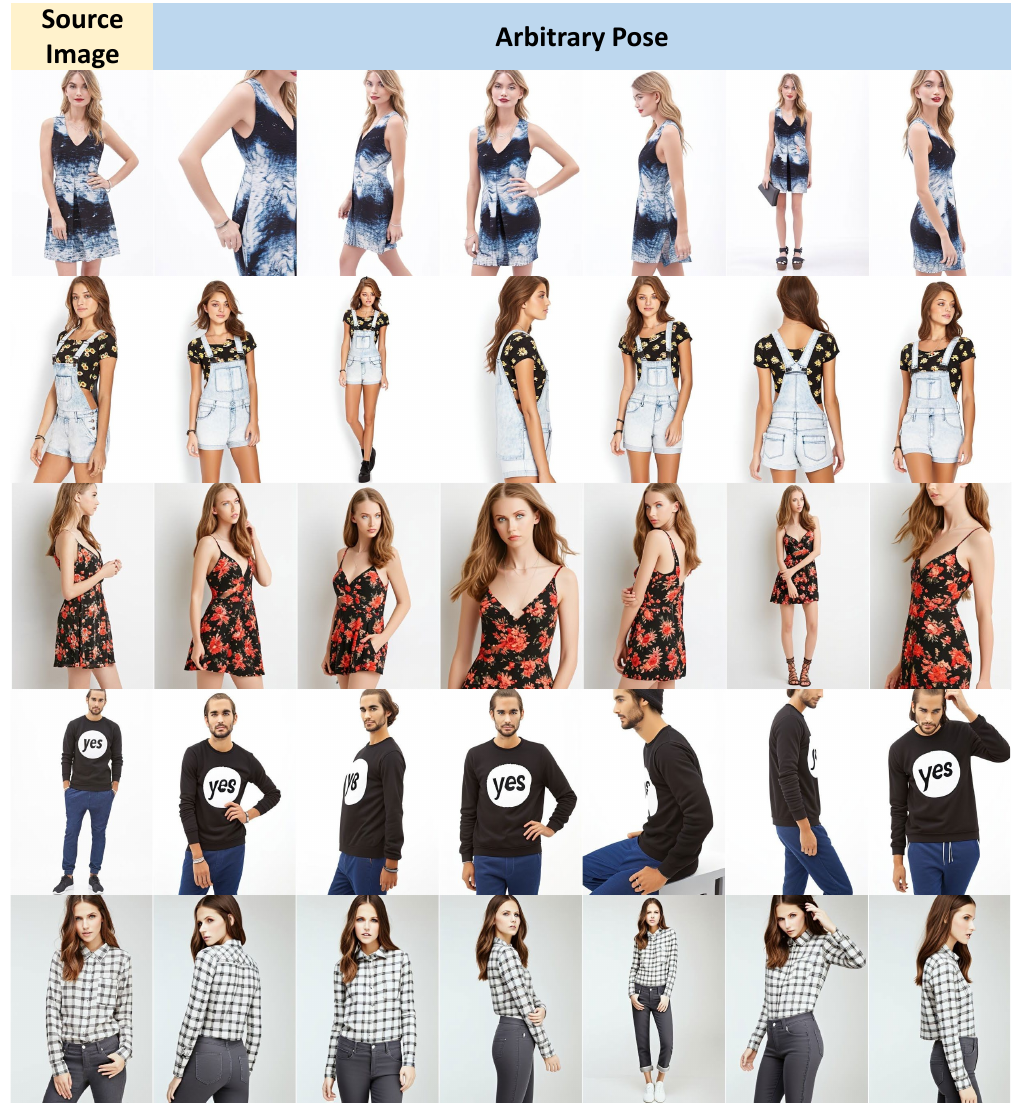}
\caption{ Additional results on arbitrary poses from the test set. }
\label{fig::supp3}
\end{figure*}

\begin{figure*}[!t]
\centering
\includegraphics[width=1\linewidth]{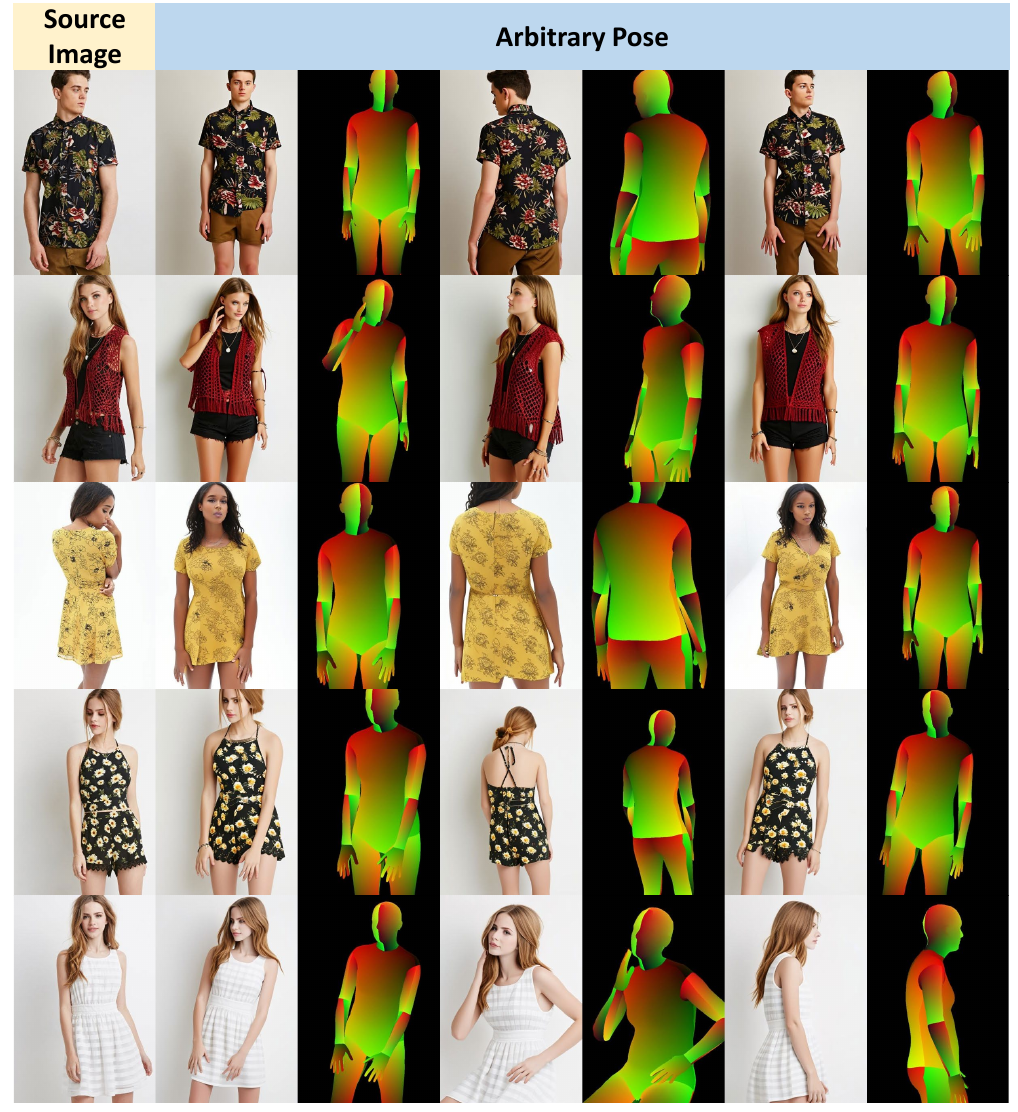}
\caption{ Additional results on rendered Densepose. The Densepose is rendered by user-defined SMPL parameters.}
\label{fig::supp4}
\end{figure*}

\begin{figure*}[!t]
\centering
\includegraphics[width=1\linewidth]{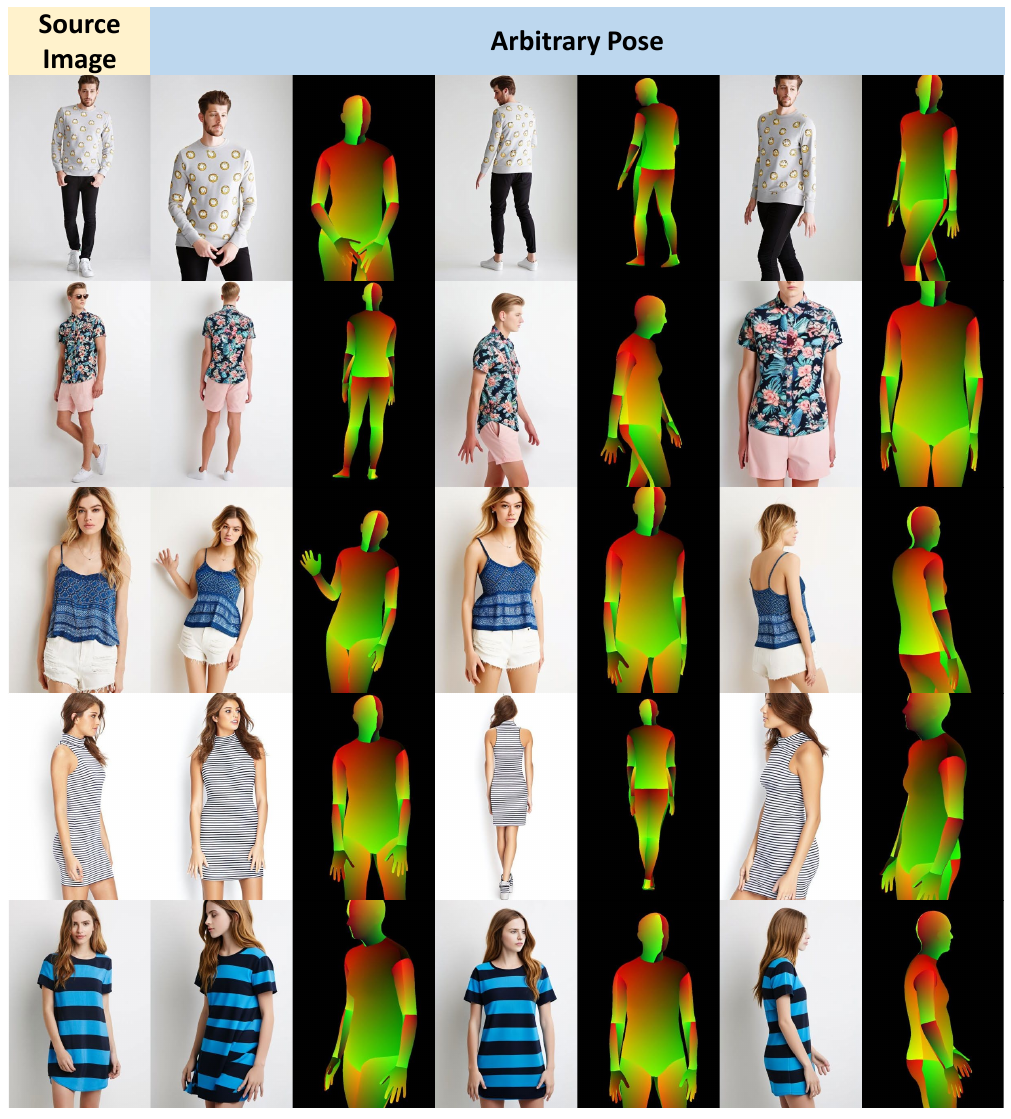}
\caption{ Additional results on rendered Densepose. The Densepose is rendered by user-defined SMPL parameters.}
\label{fig::supp5}
\end{figure*}

\noindent \textbf{Additional Qualitative Results.} We conducted two additional qualitative experiments to demonstrate the generalization ability of our method. First, we generated person images under arbitrary poses randomly selected from the test set (Fig.\ref{fig::supp1},\ref{fig::supp2},\ref{fig::supp3}), and the results show that our method consistently preserves texture patterns and person identities from the source images, even retaining complex patterns and icons, with high-quality facial details. Second, we tested the method's adaptability to user-defined poses by rendering synthetic DensePose in real-time, where synthetic poses were rendered from SMPL~\cite{SMPL} parameters estimated from the test set. The results (Fig.\ref{fig::supp4},\ref{fig::supp5}) indicate that our method can generate plausible person images with correct textures and identities. Minor weaknesses were observed in the hands and boundary regions due to differences in the generated and estimated DensePose. This problem can be mitigated through finetuning. 

\noindent \textbf{Computation Complexity of MFCA module.} Our method improves performance by introducing only 5.8\% more trainable parameters compared to baseline B1, where the MFCA modules only extend around \textbf{19M} parameters and the face projector introduces \textbf{76M} parameters. We have provided additional information , validating on A6000 GPUs. We also compare the cost with CFLD in Tab.\ref{tab::cc}.  Our method adopts a ControlNet-like structure, which minimally increases the inference cost while reducing the training time. 

\begin{table}[h]
\centering
\resizebox{\linewidth}{!}
{
\begin{tabular}{c|c|c|c}
\hline
 Methods & GPU hours (H) & inference memory (G) & inference time (s)  \\
\hline
CFLD~\cite{lu2024coarse} & 353.6 & 6.5 & 3.55 \\
B1 & 45.2 & 10.8 & 4.35 \\
MCLD (Ours)  & 54.0  & 13.9 & 4.46 \\
\hline
\end{tabular}
}
\vspace{-2mm}
\caption{Complexity Comparison of baselines.}\label{tab::cc}
\vspace{-6mm}
\end{table}

\end{document}